\documentclass[twoside,11pt]{article}
\usepackage{blindtext}

\usepackage{jmlr2e}
\usepackage{bm}
\usepackage{amsmath}
\usepackage{float}
\usepackage{algorithm}
\usepackage{algpseudocode}
\usepackage{booktabs}       
\usepackage{sidecap}
\usepackage{multirow}
\usepackage{placeins}
\usepackage{xr}

\usepackage{lastpage}
\jmlrheading{XX}{20XX}{X-X}{8/25; Revised X/XX}{X/XX}{XX-XX}{Hess et al}

\ShortHeadings{JMLR Submission}{Hess et al.}
\firstpageno{1}

\begin{document}

\title{Generating Time-Consistent Dynamics \\ with Discriminator-Guided Image Diffusion Models}
\author{\name Philipp Hess \email hess@pik-potsdam.de\\
       \name Maximilian Gelbrecht \\
       \addr Potsdam Institute for Climate Impact Research, Potsdam, Germany\\[1.0ex]
       \addr Munich Climate Center and Earth System Modelling Group,\\
       Department of Aerospace and Geodesy,\\
       School of Engineering \& Design,\\
       Technical University of Munich, Munich, Germany
       \AND
       \name Christof Schötz \\
       \name Michael Aich \\
       \name Yu Huang \\
       \name Shangshang Yang \\
       \addr Munich Climate Center and Earth System Modelling Group,\\
       Department of Aerospace and Geodesy,\\
       School of Engineering \& Design,\\
       Technical University of Munich, Munich, Germany\\[1.0ex]
       \addr Potsdam Institute for Climate Impact Research, Potsdam, Germany
       \AND
       \name Niklas Boers \\
       \addr Munich Climate Center and Earth System Modelling Group,\\
       Department of Aerospace and Geodesy,\\
       School of Engineering \& Design,\\
       Technical University of Munich, Munich, Germany\\[1.0ex]
       \addr Potsdam Institute for Climate Impact Research, Potsdam, Germany\\[1.0ex]
       \addr Global Systems Institute and Department of Mathematics,\\
       University of Exeter, Exeter, United Kingdom
       }

\editor{-}

\maketitle

\begin{abstract}%
Realistic temporal dynamics are crucial for many video generation,  processing and modelling applications, e.g., in computational fluid dynamics, weather prediction, or long-term climate simulations. 
Video diffusion models (VDMs) are the current state-of-the-art method for generating highly realistic dynamics. However, training VDMs from scratch can be challenging and requires large computational resources, limiting their wider application.
Here, we propose a time-consistency discriminator that enables pretrained image diffusion models to generate realistic spatiotemporal dynamics. The discriminator guides the sampling inference process and does not require extensions or finetuning of the image diffusion model. 
We compare our approach against a VDM trained from scratch on an idealized turbulence simulation and a real-world global precipitation dataset. Our approach performs equally well in terms of temporal consistency, shows improved uncertainty calibration and lower biases compared to the VDM, and achieves stable centennial-scale climate simulations at daily time steps.
\end{abstract}


\section{Introduction}

    Generating time-consistent sequences of images is important to many video generation and synthesis tasks \citep{ho_video_2022, ho_imagen_2022, zheng_opensora_2024, wang_zeroshot_2024, daras_warped_2024}, for example in computational fluid dynamics \citep{du_conditional_2024, li_synthetic_2024, lienen_zero_2024}, probabilistic weather forecasts \citep{price_probabilistic_2025, li_generative_2024} or climate simulations \citep{srivastava_precipitation_2024, bassetti_diffesm_2024, cachay_elucidated_2025, couairon_archesweathergen_2026}.\\
    The success of image diffusion models (IDMs) \citep{ho_denoising_2020, song_scorebased_2021} has sparked a large interest in extending their generation to time-consistent videos, achieving remarkable results \citep{ho_video_2022, ho_imagen_2022, gupta_photorealistic_2023, wang_videocomposer_2023, yang_cogvideox_2024, zheng_opensora_2024, deng_autoregressive_2025, kong_hunyuanvideo_2025}.\\
    However, training video diffusion models (VDMs) from scratch is challenging and requires large amounts of computational resources \citep{deng_efficiencyoptimized_2023}. Moreover, recent state-of-the-art VDMs are not always released open source \citep{zheng_opensora_2024}, limiting their adaptability to a wider scientific community. \\
    Therefore, efforts have been made to leverage pretrained image models for video editing tasks such as style-transfer or inpainting \citep{daras_warped_2024, chang_how_2023, ceylan_pix2video_2023, zhang_controlvideo_2023, qi_fatezero_2023}. 
    Video editing relies on full or partial temporal information in the source video that can then be combined with inference-level guidance techniques to preserve temporal consistency during the editing process.
    Such video processing tasks are also important to many scientific applications, for example, in data reconstruction using inpainting methods or downscaling applications using super-resolution techniques in fluid dynamics \citep{wan_debias_2023, bischoff_unpaired_2024}, meteorology  \citep{mardani_residual_2025} and climate science \citep{ling_diffusion_2024, plesiat_artificial_2024, bischoff_unpaired_2024, hess_fast_2025, aich_conditional_2026, furst_wind_2026, addison_machine_2026}.\\
    Generating videos with IDMs without relying on a source video or a given encoding of the dynamics is much more challenging. Most approaches rely on finetuning an IDM on video data, e.g., by inserting additional temporal layers into the architecture \citep{singer_makeavideo_2022, esser_structure_2023, wu_tuneavideo_2023,  bar-tal_lumiere_2024}, which can still be computationally demanding and requires a deep understanding of the architecture. 
    We propose a novel guidance approach, inspired by temporal discriminators in generative adversarial networks \citep{clark_adversarial_2019, ravuri_skilful_2021a, das_hybrid_2024}, for the generation of realistic, time-consistent, and stable spatiotemporal dynamics with pretrained IDMs. 
    It builds on the guidance approach of \citet{kim_refining_2023}, who use a discriminator to refine the quality of generated images with IDMs. The main distinction is that we directly train on target data only, making it computationally much cheaper, with the objective to improving temporal consistency instead of image quality.
    Our discriminator guidance is lightweight and efficient, adding only about 36\%--89\% to the generation time, and is trained independently of the IDM, making extensions of different IDMs to new downstream tasks straightforward.
    We perform a comprehensive evaluation on challenging datasets with high-dimensional chaotic dynamics, including 2D Navier-Stokes turbulence simulations and global precipitation reanalysis, using the extensive catalog of established metrics from fluid dynamics and Earth system science. 
    We find that our method performs similarly well as a VDM trained from scratch in terms of temporal dynamics, while achieving better uncertainty calibration and lower biases. Moreover, our guidance approach enables stable climate simulations over 100 years, while the VDM exhibits unstable drifts in global averages.

\paragraph{Video GANs.}
    Generative adversarial networks (GANs) have been widely explored for synthesizing temporally-consistent videos. Earlier work \citep{vondrick_generating_2016, mathieu_deep_2016} introduced the idea of using an adversarial discriminator to distinguish between real and generated video frames, which was improved in following studies \citep{saito_temporal_2017,  tulyakov_mocogan_2018, saito_train_2020}.
    DVD-GAN \citep{clark_adversarial_2019} proposed two separate discriminators for spatial and time domains, the latter being similarly motivated as our time-consistency discriminator.\\
    Video prediction GANs with temporal discriminators have shown great success in turbulence modelling \citep{xie_tempogan_2018} and probabilistic weather predictions \citep{ravuri_skilful_2021a, das_hybrid_2024}.
    However, while temporal discriminators provide powerful tools that enable the generation of dynamically consistent videos in GANs, adversarial training is generally prone to instabilities and mode collapse, making GANs challenging to optimize.
\paragraph{Video diffusion models.}
    Generative diffusion models (DMs) \citep{ho_denoising_2020, song_scorebased_2021}, have largely superseded GANs owing to their improved training stability, high-fidelity output and iterative sampling process, which enables downstream tasks without retraining \citep{song_scorebased_2021, ho_denoising_2020}.\\
    Video diffusion models \citep{ho_video_2022} have achieved state-of-the-art performance \citep{ho_imagen_2022, gupta_photorealistic_2023, wang_videocomposer_2023, yang_cogvideox_2024, zheng_opensora_2024, deng_autoregressive_2025, kong_hunyuanvideo_2025}, e.g., through latent VDMs \citep{he_latent_2023, blattmann_stable_2023, ma_latte_2024}, and improved training strategies \citep{gupta_photorealistic_2023, jin_pyramidal_2024, deng_autoregressive_2025}. Classifier-free guidance has also been explored to enable variable-length conditioning on past video frames with VDMs \citep{song_historyguided_2025}.
    Training schemes that expose the video model to corrupted context frames, such as diffusion forcing \citep{chen_diffusion_2024} and the stochastic clean context of frame-autoregressive models \citep{gu_efficiently_2022}, reduce the exposure bias in long rollouts.\\
    The ability of VDMs to model uncertainties and to produce sharp outputs makes them powerful tools, e.g., for weather prediction \citep{price_probabilistic_2025, li_generative_2024, yang_generative_2025}, super-resolution (downscaling) \citep{srivastava_precipitation_2024}, reconstructing spatiotemporal dynamics from sparse sensor measurements \citep{li_learning_2024}, emulating precipitation dynamics directly from remote sensing observations \citep{stock_diffobs_2024}, or climate model simulations \citep{bassetti_diffesm_2024, cachay_elucidated_2025, couairon_archesweathergen_2026}.\\
    However, VDMs require large computational resources for training \citep{deng_efficiencyoptimized_2023, price_probabilistic_2025}, limiting their applicability.

\paragraph{Video synthesis with image diffusion models.}
    Due to the high computational costs and lack of open source availability of VDMs, recent efforts have focused on utilizing available pretrained image diffusion models (IDMs) for video processing and editing tasks.
    In video processing tasks, the temporal dynamics are usually given in a source video that needs to be transformed in a time-consistent manner, for applications such as style-transfer, inpainting, or super-resolution \citep{sun_diffusion_2026}. Approaches to preserve temporal consistency include correlated (warped) noise \citep{daras_warped_2024, chang_how_2023}, or transitioning from spatial to temporal-attention blocks \citep{ceylan_pix2video_2023, zhang_controlvideo_2023, qi_fatezero_2023, geyer_tokenflow_2023, khachatryan_text2videozero_2023, wang_zeroshot_2024}.\\
    IDMs have been adapted to applications in fluid dynamics, weather prediction and climate modelling. 
    Some applications take temporal dynamics explicitly into account, e.g., in data-assimilation \citep{rozet_scorebased_2023} and spatio-temporal downscaling \citep{srivastava_precipitation_2024, schmidt_spatiotemporally_2025}. Sampling guidance from a numerical weather prediction model has also been used to improve the weather forecast from a VDM  \citep{hua_weather_2024}. Further, IDMs have been combined with a deterministic forecast neural network to produce dynamically consistent simulations from weather to climate time scales \citep{yang_generative_2025}.\\
    Many applications, however, employ IDMs to process dynamical simulations in each time step without taking time consistency into account, e.g., for downscaling (super-resolution) climate simulations \citep{bischoff_unpaired_2024, hess_fast_2025, aich_conditional_2026, mardani_residual_2025, ling_diffusion_2024, addison_machine_2026}, data-assimilation \citep{huang_diffda_2024a}, or data reconstruction \citep{plesiat_artificial_2024}, which could potentially be improved with our method.\\
    When generating videos with IDMs, e.g., from a given starting frame, most work relies on finetuning a pretrained model by inserting temporal-attention layers \citep{singer_makeavideo_2022, esser_structure_2023, wu_tuneavideo_2023,  bar-tal_lumiere_2024}. Notably, \citep{blattmann_align_2023} use a temporal discriminator for finetuning the extended IDM. Our approach, in contrast, is agnostic of the IDM architecture and does not require finetuning.

\paragraph{Discriminator guidance.}
    Inspired by GANs, discriminators have been employed during diffusion model training to improve the performance \citep{wang_diffusiongan_2022, huang_diffdis_2023, ko_adversarial_2024} or enhance sampling speed with adversarial distillation \citep{sauer_adversarial_2024, yin_improved_2024}.
    Discriminators have also been proposed as purely inference-level guidance to improve the image quality of IDMs \citep{kim_refining_2023}, to pair separately trained video and audio diffusion models \citep{hayakawa_discriminatorguided_2024}, or to generate molecular graphs \citep{kelvinius_discriminator_2024, kerby_trainingfree_2024}.\\
    Similarly, our time-consistency discriminator is only applied during inference as guidance. A notable difference to \citep{kim_refining_2023, hayakawa_discriminatorguided_2024}, is that our training does not require samples generated with an IDM, which can be computationally costly. 

    \begin{figure}
        \centering
        \includegraphics[width=1.0\linewidth]{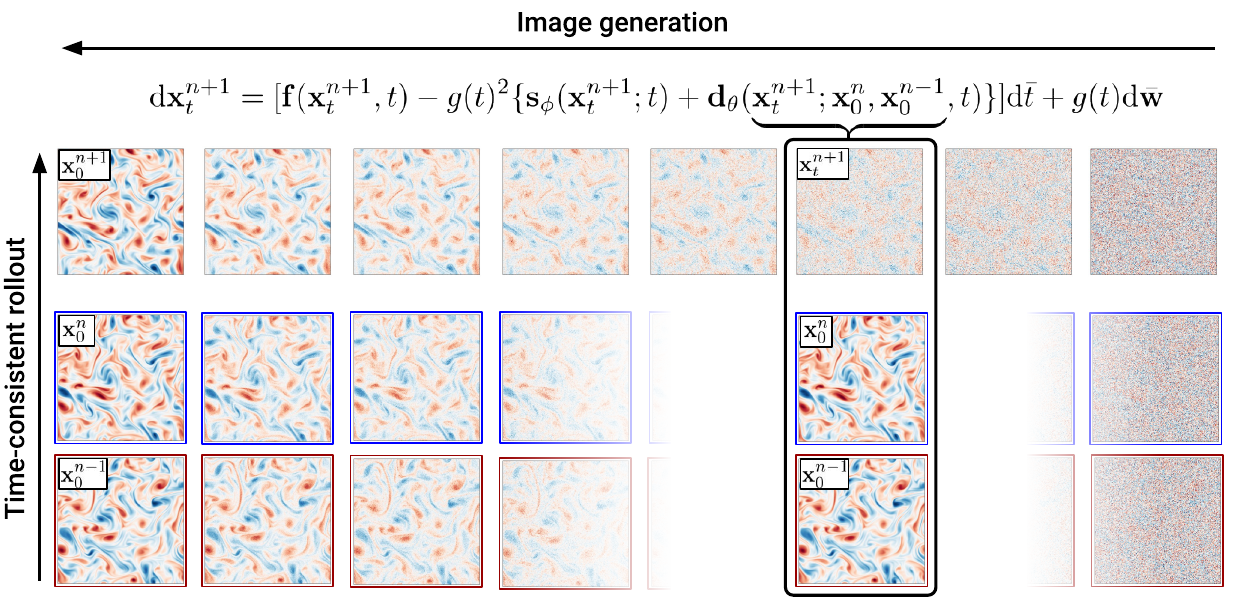}
	    \caption{\textbf{Overview sketch} of the time-consistency discriminator guidance for generating images in a dynamically realistic sequence. The discriminator guidance $\bm{d}_{\theta}(\cdot)$ uses the current and past time frames, $\bm{x}^{n}$ and $\bm{x}^{n-1}$, to guide the denoising generation of the next $\bm{x}^{n+1}$.}
        \label{fig:method_overview}
    \end{figure}

\section{Methods}
\label{sec:guidance}

\paragraph{Diffusion models.}
    Diffusion models \citep{ho_denoising_2020, song_scorebased_2021} learn to generate data from a target distribution $\bm{x}_{0} \sim p_{\textrm{data}}(\bm{x})$ with a time-reversed denoising process that starts with an initial noise sample, e.g. Gaussian white noise $\bm{x}_{T} \sim \mathcal{N}(\bm{0}, \sigma_{\text{max}}^2\mathbf{I})$, and can be formulated as a reversed stochastic differential equation (SDE) \citep{anderson_reversetime_1982},  
    \begin{equation}
        \mathrm{d}\bm{x}_t = [ \bm{f}(\bm{x}_t, t) - g(t)^2 \nabla_{\bm{x}_t} \log p_{t}(\bm{x}_t) ] \mathrm{d}\bar{t} + g(t)\mathrm{d}\bm{\bar{w}},
         \label{eq:reverse_sde}
    \end{equation}
    where $\bm{x}_t \in \mathbb{R}^{B \times C \times H \times W}$ is a noised image
    of batch size $B$ with $C$ channels and $H$, $W$ pixels in height and width dimension, respectively. The drift term is given by $\bm{f}(\cdot)$, $\mathrm{d}\bm{\bar{w}}$ adds Wiener noise where the bar denotes a time reversal, and $\nabla_{\bm{x}_t} \log p_{t}(\bm{x}_t)$ is the score function of the noised target distribution. The noise strength $\sigma_t$ with lower and upper bounds  $\sigma_{\text{min}}$ and $\sigma_{\text{max}}$, respectively, decreases during the reverse processes, following a prescribed schedule $g(t)$ (e.g. see Eq.~\ref{eq:noise_schedule}). The score function in Eq.~\ref{eq:reverse_sde} is typically intractable but can be learned with a neural network  $\bm{S}_\phi(\cdot)$ \citep{karras_elucidating_2022},   
    \begin{equation}
        \nabla_{\bm{x}_t} \log p_{t}(\bm{x}_t) \approx \frac{\bm{S}_\phi(\bm{x}_t; t) - \bm{x}_t}{\sigma_t^2} =: \bm{s}_{\phi}(\bm{x}_t; t),
        \label{eq:unconditional_score}
    \end{equation}
     using the loss function $\mathcal{L}(\phi) = \mathbb{E}_{\bm{x} \sim p_{\textrm{data}}(\bm{x}), \bm{\epsilon}_t \sim \mathcal{N} (\mathbf{0}, \sigma^2_t\mathbf{I})} [ w(t) \Vert \bm{S}_\phi(\bm{x}_0 + \bm{\epsilon}_t;t) - \bm{x}_0 \Vert ^2_2 ]$, where $w(t)$ is a weighting function (see details in Appendix \ref{app:diffusion_model}).

\paragraph{Time-consistency guidance.}
    We propose a discriminator that guides the reverse diffusion process in Eq.~\ref{eq:reverse_sde} to generate time-consistent sequences of images (see Fig.~\ref{fig:method_overview}).
    A discrete, temporally ordered time series of $N$ images is denoted as $\{\bm{x}^n_t | n=1,2,...,N\}$, where the superscript represents the physical time step and the subscript the diffusion time. We train a discriminator to distinguish between images that sampled in a time-consistent manner, i.e. that are conditioned on a noise-free ($t=0)$ sequence of the current and $m$ previous time steps, $\{\bm{x}^{n-m}_0,\bm{x}^{n-m+1}_0, ..., \bm{x}^{n}_0\} = \bm{x}^{(n-m):n}_0$, following $\bm{x}^{n+1}_t \sim p_t (\bm{x}^{n+1}_t|\bm{x}^{(n-m):n}_0)$, and random samples without temporal ordering $\bm{x}^{n+1}_t \sim p_t(\bm{x}^{n+1}_t)$. 
    A optimal discriminator has then the form \citep{goodfellow_generative_2014, kim_refining_2023}, 
    \begin{equation}
	    D_{\theta} (\bm{x}^{n+1}_t; \bm{x}^{(n-m):n}_0, t ) = \frac{p (\bm{x}^{n+1}_t|\bm{x}^{(n-m):n}_0 )}{p (\bm{x}^{n+1}_t|\bm{x}^{(n-m):n}_0) + p \left(\bm{x}^{n+1}_t \right)}.
	\label{eq:optimal_discriminator}
    \end{equation}
    Computing the scores, by applying the logarithm and gradient, on both sides of Eq.~\ref{eq:optimal_discriminator},
    \begin{equation}
    \nabla_{\bm{x}^{n+1}_t} \log \left(\frac{D_{\theta}(\bm{x}^{n+1}_t; \bm{x}^{(n-m):n}_0, t)}{1-D_{\theta}(\bm{x}^{n+1}_t; \bm{x}^{(n-m):n}_0, t)} \right) =  \nabla_{\bm{x}^{n+1}_t} \log \left(\frac{p(\bm{x}^{n+1}_t|\bm{x}^{(n-m):n}_0)}{p(\bm{x}^{n+1}_t)} \right),
    \label{eq:log_ratio}
    \end{equation}
    allows us to define the time-consistency guidance as
    \begin{equation}
	    \bm{d}_\theta(\bm{x}^{n+1}_t; \bm{x}^{(n-m):n}_0,t) := \nabla_{\bm{x}^{n+1}_t} \log \left(\frac{D_{\theta}(\bm{x}^{n+1}_t; \bm{x}^{(n-m):n}_0, t)}{1-D_{\theta}(\bm{x}^{n+1}_t; \bm{x}^{(n-m):n}_0, t)} \right).
    \label{eq:discriminator_guidance}
    \end{equation}
    The guidance term in Eq.~\ref{eq:discriminator_guidance} can then be added to the unconditional score to enable time-consistent sampling of images in an autoregressive manner using the reverse SDE in Eq.~\ref{eq:reverse_sde},
    \begin{equation}
    \mathrm{d}\bm{x}^{n+1}_t = [ \bm{f}(\bm{x}^{n+1}_t, t) - g(t)^2 \{\bm{s}_{\phi}(\bm{x}^{n+1}_t;t) + \lambda_t \bm{d}_\theta(\bm{x}^{n+1}_t; \bm{x}^{(n-m):n}_0, t) \} ] \mathrm{d}\bar{t} + g(t)\mathrm{d}\bm{\bar{w}} 
    \label{eq:guided_sde}
    \end{equation}
    where the strength of the guidance is controlled through the parameter $\lambda_t$ (see Appendix \ref{app:derivaion} for a more detailed discussion).
   In our experiments we condition the discriminator on the current and the previous time step ($m = 1$). Two frames are the smallest context that determines a tendency: a single frame fixes the state but not its direction of change, so a discriminator conditioned on one frame can judge only how far a candidate frame is from the current one, not whether it continues the evolution, and the temporal-order negatives of Eq.~\ref{eq:loss} would then be distinguishable only by the magnitude of the change.
   With two frames, the first-order time derivative is defined and the candidate can be judged against it, which is the information a multistep integrator uses and the conditioning adopted by ML weather models \citep{price_probabilistic_2025}. 
   We use the same context for our video DM baseline, so that the comparison between the two is not confounded by the context length.
    
\paragraph{Sample-based and sample-free guidance.}
    We train the discriminator (``sample-free'') on target data only to distinguish between random fields from $p_{\text{data}}(\bm{x}^{n+1})$  and temporally ordered (consistent) ones from $p_{\text{data}}(\bm{x}^{n+1} | \bm{x}^{(n-m):n})$. Hence, our approach does not require samples from the diffusion model as in \citep{kim_refining_2023} (called ``sample-based'' hereafter).
    To illustrate the difference we can decompose the log-likelihood ratio used for guidance in the following way, where we denote the image diffusion model distribution with $p_{\text{IDM}}$,
    \begin{equation}
    \log \left(\frac{p_{\text{data}}(\bm{x}^{n+1} | \bm{x}^{(n-m):n})}{p_{\text{IDM}}(\bm{x}^{n+1})} \right) = \underbrace{\log \left(\frac{p_{\text{data}}(\bm{x}^{n+1} | \bm{x}^{(n-m):n})}{p_{\text{data}}(\bm{x}^{n+1})} \right)} _{\text{temporal consistency}} + \underbrace{\log \left(\frac{p_{\text{data}}(\bm{x}^{n+1})}{p_{\text{IDM}}(\bm{x}^{n+1})}\right)}_{\text{IDM error}}.
    \label{eq:log_ration_extension}
    \end{equation}
    From Eq.~\ref{eq:log_ration_extension} we see that a discriminator trained to distinguish temporally ordered fields from the target dataset and random fields from an IDM would combine two objectives. Namely temporal consistency (first term on the RH side) and the marginal mismatch between IDM and target data.
    Hence it can achieve high accuracy by focusing on the second objective while disregarding temporal consistency.
    When the IDM and target marginal distributions agree with high accuracy, however, it becomes easier for the discriminator to achieve high score by detecting temporal ordering.
    Our ``sample-free'' approach thus assumes that the IDM and target marginals will closely agree and therefore does not require the costly generation of random IDM samples.

\begin{figure}
        \centering
        \includegraphics[width=0.60\linewidth]{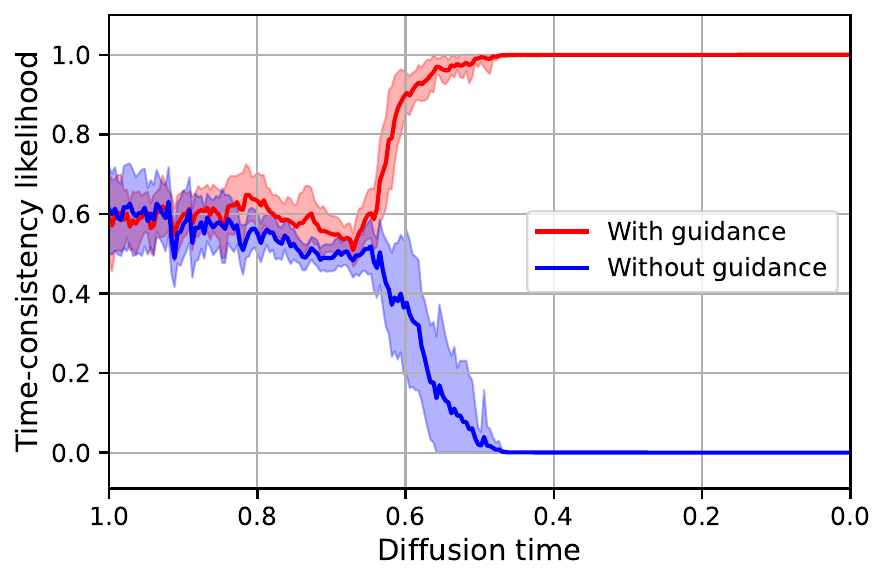}
	    \caption{\textbf{Time-consistency prediction} of the discriminator network during sampling of vorticity fields with guidance switched on (red) or off (blue). The mean over 50 samples is given by the solid line, and the shaded area shows the standard deviation. With decreasing noise scales in the reserve diffusion process ($t_{\max}=1 \rightarrow t_{\min}=0$), the discriminator network reliably predicts whether samples are time-consistent or not. }
        \label{fig:discriminator_likelihood}
    \end{figure}

\paragraph{Discriminator training.}
    The discriminator is trained as a binary classifier $D_\theta$ mapping $(\bm{x}^k_t; \bm{x}^{(n-m):n}_0, t) \mapsto q$, conditioned on previous, denoised time frames $\bm{x}^{(n-m):n}_0$, and the diffusion noise time $t$, to predict the probability $q$ of a noised image $\bm{x}^k_t = \bm{x}^k_0 + \bm{\epsilon}_t$, $\bm{\epsilon}_t \sim \mathcal{N}(\bm{0}, \sigma^2_t\bm{1})$, being temporally consistent with the current and $m$ previous time frames, i.e., whether $\bm{x}^k_t = \bm{x}^{n+1}_t$. The same noise schedule as for training the diffusion model is used for $\sigma_t$ (see Appendix \ref{app:diffusion_model}). 
    We use the standard cross entropy loss as a training objective \citep{kim_refining_2023},
    \begin{equation}
    \mathcal{L}_{CE} (\theta) = - \mathbb{E}_{n,l} \left[ \log D_{\theta}(\bm{x}^{n+1}_t;\bm{x}^{(n-m):n}_0,t)   + \log ( 1-D_{\theta}(\bm{x}^{n+l}_t; \bm{x}^{(n-m):n}_0, t)) \right],
    \label{eq:loss}
    \end{equation}
    with $\mathbb{E}_{n,l} := \mathbb{E}_{ n \sim \mathcal{U} (1, N), l \sim \mathcal{N}_{\mathbb{Z}}(\mu,\sigma_{\text{step}}^2) \setminus \{ 1 \} }$, where we uniformly sample a time step $n$ from the dataset of $N$ samples, and introduce an importance sampling for non-time consistent samples ($l\neq 1$) to prioritize time steps from the vicinity of the next time step, $k=n+1$, using a normal distribution of integers $l \sim \mathcal{N}_{\mathbb{Z}}(\mu,\sigma_{\text{step}}^2) \setminus \{ 1 \}$, and we set $\mu=1$, $\sigma_{\text{step}}=2$. The motivation is that fields close to the next time step ahead are hardest to distinguish for the network, due to their high correlation.
    We apply random cropping of the fields during training of the discriminator and evaluate it on full fields during inference (see appendix \ref{app:discriminator_training} for details). 
    The cropping was adapted to force the discriminator to focus on different spatial scales, instead of only learning large scale dynamics in global fields. 
    Fig.~\ref{fig:discriminator_likelihood} shows the time consistency prediction of the trained discriminator network during inference. See Appendix \ref{app:discriminator_training} for training and architecture details.

\section{Experiments}

\paragraph{Data.}
\label{subsec:data}
    We evaluate our method on two challenging datasets: an idealized fluid dynamical Navier-Stokes simulation and real-world observational precipitation data from the ERA5 reanalysis \citep{hersbach_era5_2020}.\\
    A two-dimensional vorticity simulation is performed by numerically solving the Navier-Stokes equation in vorticity formulation with a 4th-order Runge-Kutta solver on a $256 \times 256$ grid with periodic boundary conditions and stochastic forcing. We use 47k samples for training and 13k for validation and test set, respectively (see Appendix \ref{app:navier_stokes_data} for details). The time-consistency evaluation in the following is performed over 4k samples.\\
    For the second experiment, we use global precipitation fields from the ERA5 reanalysis, which combines high-resolution numerical weather model simulations with different observational sources using data assimilation (see details in Appendix \ref{app:precipitation_data}), as a complementary and real-world dataset that is challenging due to its spatiotemporal intermittency, inherent stochasticity, and skewed distributions. 
    The horizontal spatial resolution is $1^\circ$ degree, which corresponds to $180 \times 360$ grid cells in latitude (height) and longitude (width) direction, respectively. We split the daily data into periods of 1979-2000, 2001-2010 and 2011-2020 for training, validation and testing.
    
\paragraph{Baselines.}
    We compare our time-consistency discriminator guidance method to sampling from the unconditional DM without guidance, and a video DM baseline trained from scratch. The video DM is set up in an autoregressive manner \citep{ho_video_2022, price_probabilistic_2025}, with a conditional score network $s_\psi(\bm{x}^{n+1}_t;\bm{x}^{n}_0,\bm{x}^{n-1}_0,t) \approx \nabla_{\bm{x}^{n+1}_t} \log p_t(\bm{x}^{n+1}_t|\bm{x}^{n}_0, \bm{x}^{n-1}_0) $. We find that the same hyperparameters work well for both the video DM and unconditional DMs (see Appendix \ref{app:diffusion_model} for details).
    We further compare against a conditional GAN with temporal and spatial discriminators in the design of DGMR \citep{ravuri_skilful_2021a}, whose generator is the video DM
    U-Net trained adversarially instead of by denoising. Architecture, training and
    checkpoint selection are given in Appendix~\ref{app:gan}.

\paragraph{Sampling.}
    We use the stochastic EDM sampler \citep{karras_elucidating_2022} for the unconditional and video DMs, with the same parameters for both models (see Tab.~\ref{tab:diffusion_config}).  We apply the discriminator guidance Eq.~\ref{eq:guided_sde} in both the first and second-order solver steps (see algorithm \ref{alg:sampler}), which we find to be important to achieve good performance. 

\section{Results}
    We evaluate our time-consistency guidance approach against the baselines on Navier-Stokes fluid dynamical simulations of turbulence, and observational daily precipitation from state-of-the-art reanalysis data (ERA5) \citep{hersbach_era5_2020}, using established metrics in dynamical systems theory and Earth system science (see Appendix \ref{app:metrics} for definitions).
\begin{figure}
    \centering
    \includegraphics[width=0.96\linewidth]{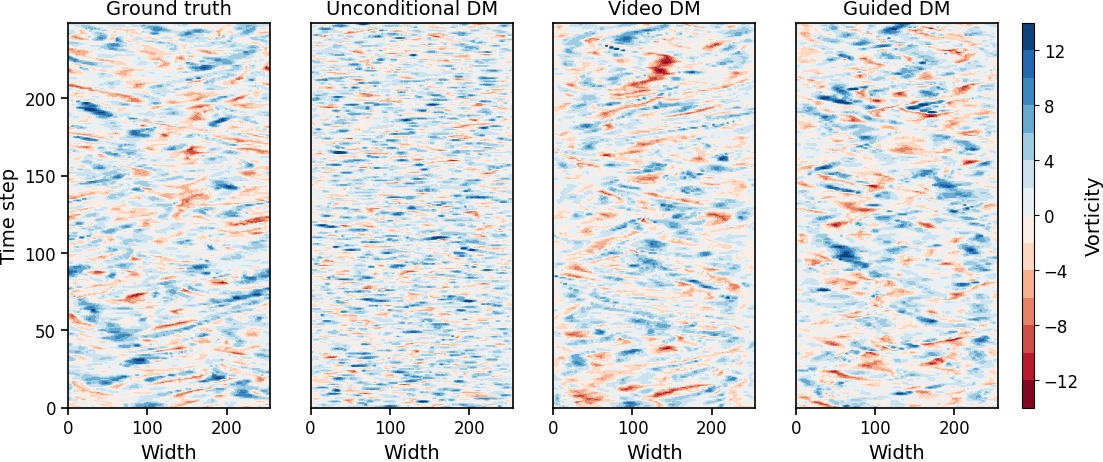}
    \caption{\textbf{Hovmöller diagrams}, often used to visualize spatiotemporal dynamics
    and, in particular, the propagation of waves in fluid dynamics and meteorology, 
    are computed for the 2D vorticity simulation as the mean over a vertical band of grid columns for (from left to right) the ground truth numerical simulation, the unconditional DM, the video DM, and our guidance approach. 
    The guidance method and video DM generate dynamics indistinguishable from the ground truth.
    }
    \label{fig:vorticity_hovmoeller}
\end{figure}
\paragraph{2D Navier-Stokes turbulence.}
    A qualitative comparison of samples is shown in Fig.~\ref{fig:voriticity_samples} for the first five and last time frames of the ground truth and generated simulations. A video of the generated dynamics is also provided\footnote{\url{https://youtu.be/JMwsZi_b-uk}}.
    While the pairing between generated and ground truth samples is quickly lost due to the chaotic non-linear dynamics, both video DM and discriminator-guided DM produce realistic dynamics in contrast to the unconditional DM. All generative DMs remain sharp over the entire 4000-step rollout. \\
    To better visualize the dynamics, we compute Hovmöller diagrams \citep{hovmoller_troughandridge_1949}, showing the average over a vertical band of 10 grid columns in the center of the fields (Fig.~\ref{fig:vorticity_hovmoeller}). 
    Our guidance approach is able to reproduce the elongated wave-like structures over multiple time steps that can be seen in the ground truth, video DM and guided DM simulations, but are absent in the unconditional DM output.\\
    We quantify the similarity in the dynamics seen in Fig.~\ref{fig:vorticity_hovmoeller} by computing the Wasserstein-1 distance between consecutive rows in the Hovmöller diagram and compare their distributions in Fig.~\ref{fig:vorticity_statistics}a and errors in Fig.~\ref{fig:vorticity_stats_error}a. A close match between the distributions of the ground truth, video and guided DM output can be seen.
 \begin{figure}[h]
    \centering
    \includegraphics[width=0.9\linewidth]{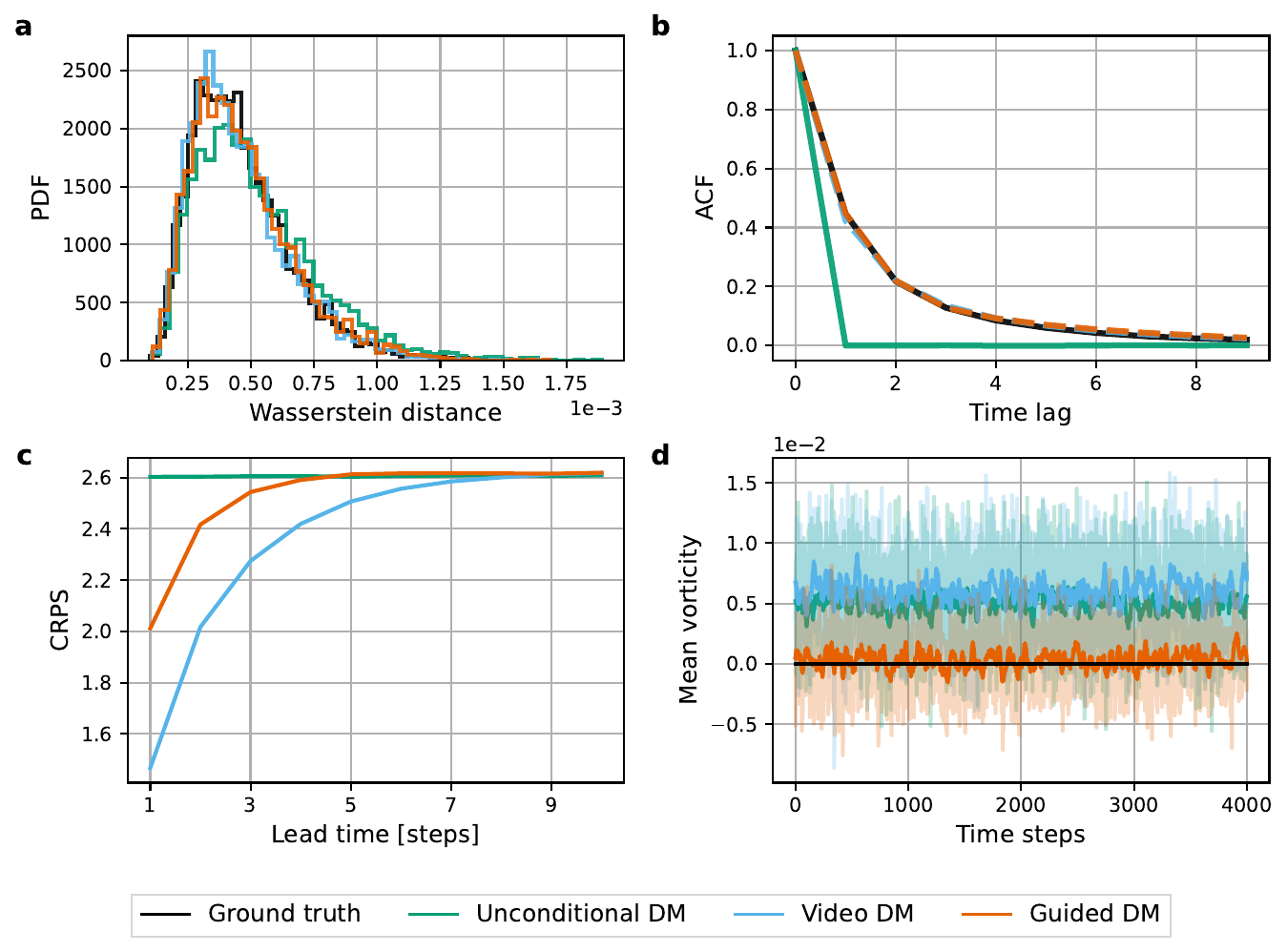}
    \caption{\textbf{Quantitative evaluation of 2D Navier-Stokes turbulent vorticity dynamics} in terms of Wasserstein distances between consecutive rows of the Hovmöller diagram in Fig.~\ref{fig:vorticity_hovmoeller} (a), autocorrelation function (ACF) (b), continuous ranked probability score (CRPS) (c) and running window spatial mean as solid line with the actual time series shown as shades (d), for the ground truth simulation (black), the unconditional DM (green), video DM (blue) and the guided DM (red). Note that only the guided DM achieves an unbiased representation of the vorticity. 
    The corresponding evaluation of the GAN baseline is shown in Appendix Fig.~\ref{fig:supplement_gan_vorticity_stats}.
    }
    \label{fig:vorticity_statistics}
\end{figure}
    We compute autocorrelation functions (ACFs) with a time lag of up to 10 time steps (Fig.~\ref{fig:vorticity_statistics}b, Fig.~\ref{fig:vorticity_stats_error}b). Both the video and guided DM achieve very accurate ACFs that are indistinguishable from the ground truth, whereas the unconditional DM generates uncorrelated samples as expected.\\
    Forecast skill is compared in terms of the continuous ranked probability score (CRPS) \citep{gneiting_probabilistic_2007} using a 50-member ensemble, 10-step lead times, and 100 forecasts (Fig.~\ref{fig:vorticity_statistics}c). We find that the video DM outperforms the guided DM in terms of forecast skill for the first six lead times, after which both converge to the climatological error. We compute the spread skill ratio and find that the guided DM shows a better calibration over all lead times (Fig.~\ref{fig:vorticity_ssr}).
    The spread-skill ratio of the video DM rises from 0.87 at the first lead to 0.96 at lead ten, that of the guided DM stays at 0.97--1.02 throughout (Fig.~\ref{fig:calibration_vorticity}). 
    Decomposing the ratio shows that the video DM has the lower error at short leads but a spread well below it, with U-shaped rank histograms, i.e.\ it is sharper but under-dispersive, whereas the guided DM's spread matches its error from the first lead on and its rank histograms are close to flat.\\
    In terms of global mean vorticity, both the unconditional and video DM show significant biases. The guidance method, in contrast, achieves a substantially lower bias (Fig.~\ref{fig:vorticity_statistics}d, Fig.~\ref{fig:vorticity_stats_error}c).\\
     The GAN baseline (Appendix~\ref{app:gan}) is stable over 4000 steps, but generates sheared, diagonally aligned filaments instead of coherent vortices (Fig.~\ref{fig:supplement_gan_vorticity_samples}) and a quasi-stationary large-scale pattern without propagating features (Fig.~\ref{fig:supplement_gan_vorticity_hovmoeller}).
    While it matches the Wasserstein distances, it decorrelates too fast (lag-one ACF 0.22 vs.\ 0.45) and has a constant mean offset of about 0.15 (Fig.~\ref{fig:supplement_gan_vorticity_stats}, \ref{fig:supplement_gan_vorticity_stats_error}).
    Its CRPS is 8--11\% worse than that of the guided DM, and its spread-skill ratio drops from 0.98 to 0.83 over ten leads (Fig.~\ref{fig:supplement_gan_vorticity_spread_skill_ratio}).
\begin{figure}[h]
    \centering
    \includegraphics[width=0.92\linewidth]{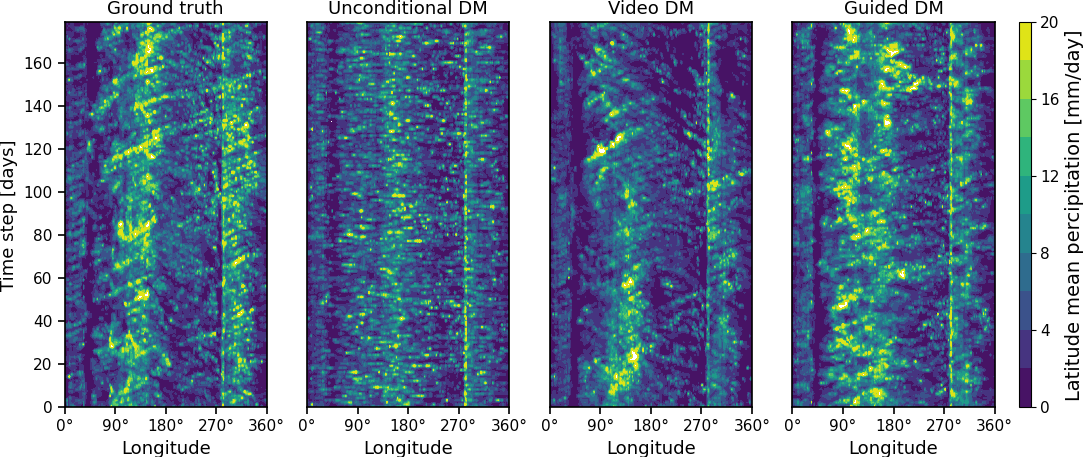}
    \caption{\textbf{Hovmöller diagrams} of the global daily precipitation simulation (from left to right) are computed for 180 days as a mean over the latitude band from $10^\circ$S to $10^\circ$N for the ground truth ERA5, unconditional DM, video DM, and our guidance approach.}
    \label{fig:precipitation_hovmoeller}
\end{figure}

\paragraph{ERA5 global precipitation.}

\begin{figure}[h]
    \centering
    \includegraphics[width=0.95\linewidth]{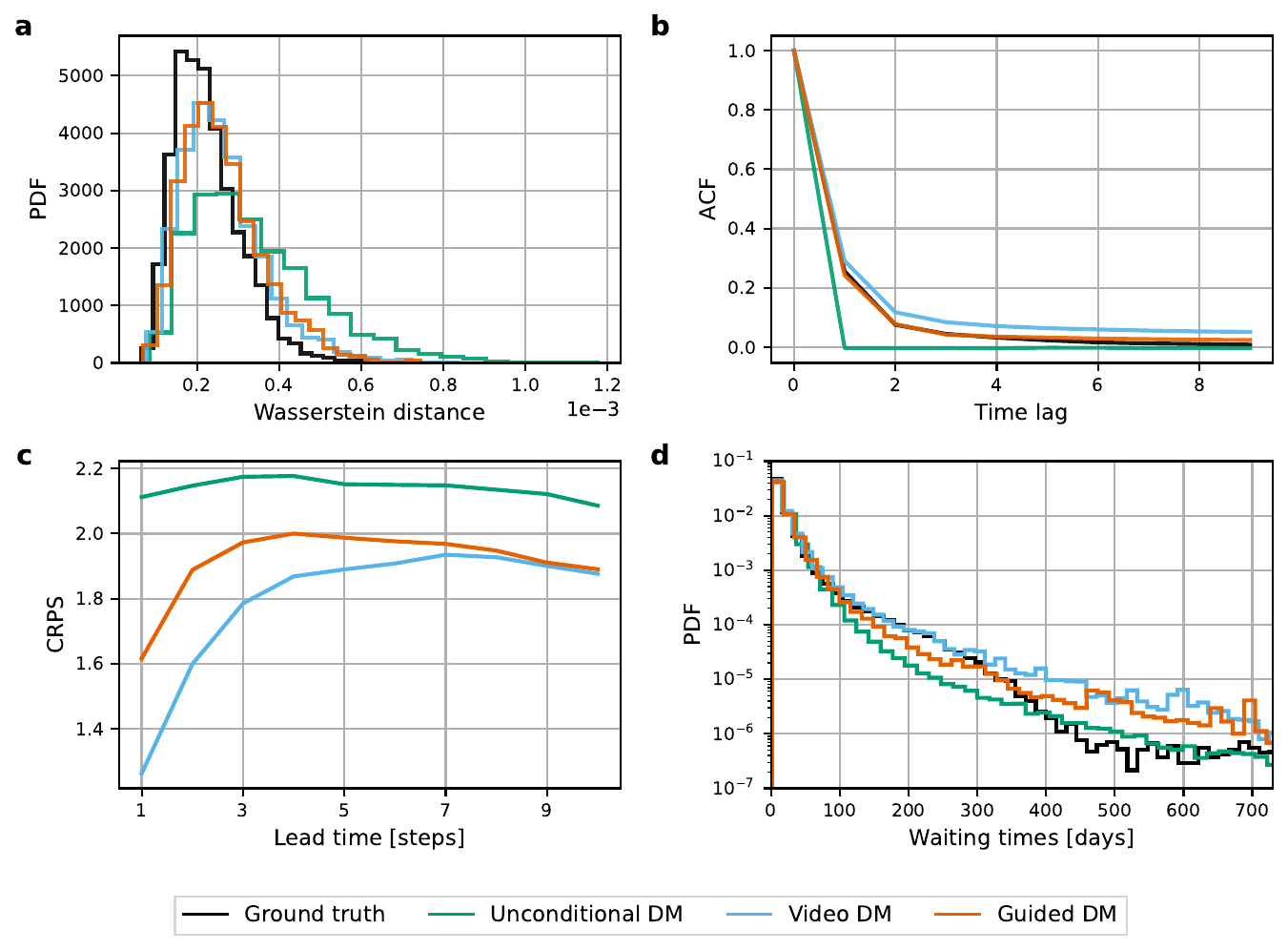}
    \caption{\textbf{Quantitative evaluation of daily precipitation dynamics} in terms of Wasserstein distances between consecutive rows of the Hovmöller diagram in Fig.~\ref{fig:precipitation_hovmoeller} (a), autocorrelation functions (ACFs) (b), CRPS forecast skill (c), extreme event waiting time distributions (d), for the ground truth ERA5 (black), the unconditional DM (green), video DM (blue) and our guidance approach (red).
    The corresponding evaluation of the GAN baseline is shown in Appendix Fig.~\ref{fig:supplement_gan_precipitation_stats}.}
    \label{fig:precipitation_statistics}
\end{figure}

\begin{figure}[h]
    \centering
    \includegraphics[width=0.85\linewidth]{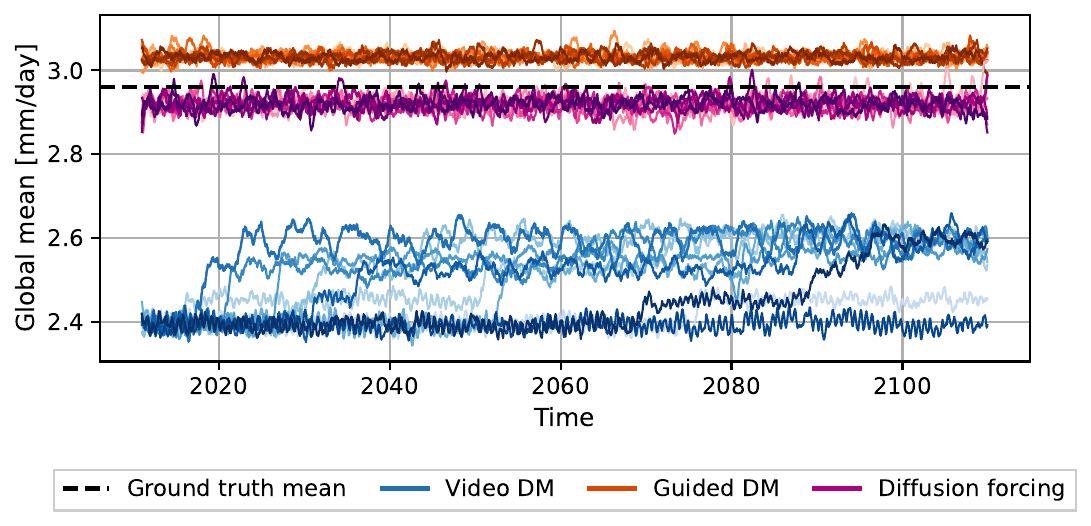}
    \caption{\textbf{Long-term precipitation simulations} are shown as an annual rolling global mean for the video DM (blue), diffusion-forcing DM (purple), our guidance method (red). The ERA5 mean is shown for reference (black dashed).
    Shadings of one color denote different ensemble members, showing that the video DM exhibits randomly occurring drifts, whereas the diffusion forcing and guided DM remain stable. 
    }
    \label{fig:precipitation_100_years_forcing}
\end{figure}

\begin{figure}[h]
    \centering
    \includegraphics[width=1.0\linewidth]{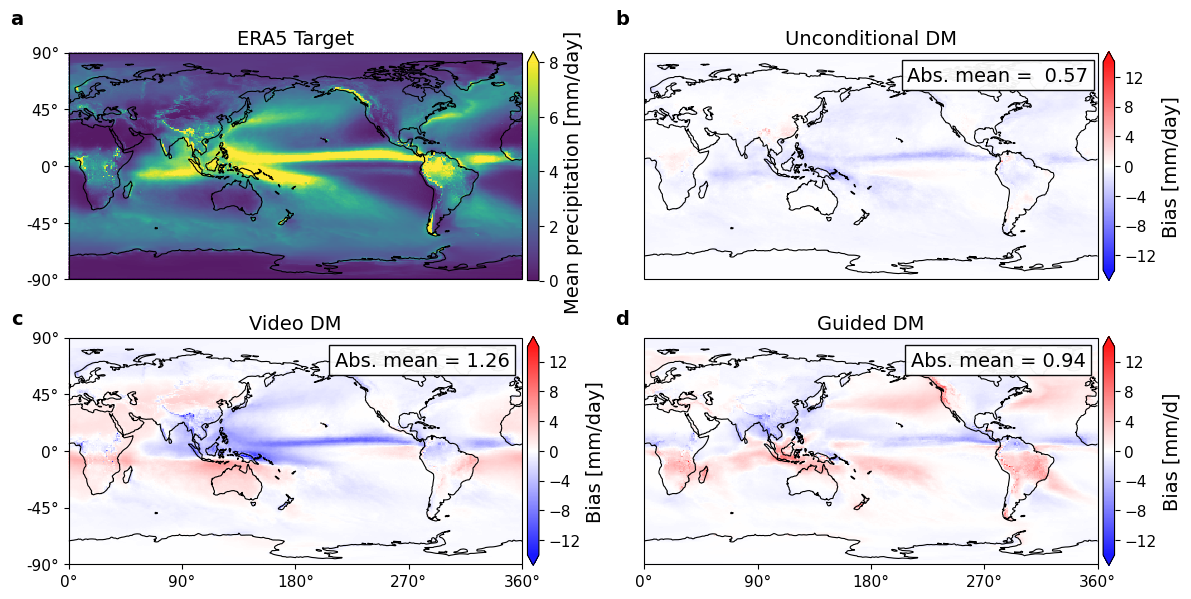}
    \caption{\textbf{Global mean bias} (see Appendix \ref{app:metrics} for definition) comparison showing, (a) the test set mean of the ERA5 ground truth, (b) the bias of the unconditional DM, (c) the video DM, (d) our guided DM. Mean absolute bias values are given in the top right.}
    \label{fig:precipitation_bias}
\end{figure}

    The first two and last daily precipitation fields from the ground truth test set and generated time series are shown in Fig.~\ref{fig:precipitation_samples}, videos of the simulations are also provided\footnote{\url{https://youtu.be/noRxrb9trpQ}}.
    All DMs produce realistic spatial patterns and remain sharp for the entire 10-year rollout, while the video DM and guided DM show qualitatively much more realistic dynamics than the unconditional DM.\\ 
    The generated dynamics are again compared with Hovmöller diagrams over 180 days, using a latitude band from $10^\circ$S to $10^\circ$N. The unconditional DM produces visible random patterns that are distinctively different from the ERA5 target data (Fig.~\ref{fig:precipitation_hovmoeller}). Our guidance approach enables the unconditional DM to produce realistic dynamics, similar to the video DM and target data, with characteristic west-to-east wave-like patterns, which are challenging to capture even for state-of-the-art climate models \citep{ahn_mjo_2020}.\\ 
    We use the Wasserstein-1 distance again to quantify the similarity in the dynamics seen in the Hovmöller diagram in Fig.~\ref{fig:precipitation_statistics}a (see errors in Fig.~\ref{fig:precipitation_stats_error}a). We find that the unconditional DM produces a flat distribution with larger distances shifted to the right of the target data distribution, which is narrower. Both the guided and the video DM capture the target distribution more accurately. 
    Our guidance method produces a very accurate autocorrelation function, slightly outperforming the video DM for longer lags (Fig.~\ref{fig:precipitation_statistics}b, Fig.~\ref{fig:precipitation_stats_error}b).
    We again perform 100 ensemble forecasts with an ensemble size of 50 members and compute the CRPS to evaluate the skill (Fig.~\ref{fig:precipitation_statistics}c). We find that the video DM has a better forecast skill than the guided DM for lead times up to eight days, after which both converge. 
    The guided DM is also the better calibrated model for precipitation, with a spread-skill ratio of 0.93--0.99 against 0.82--0.87 for the video DM    (Fig.~\ref{fig:precipitation_ssr} and \ref{fig:calibration_precipitation}). 
    The rank histograms show that the video DM is not only under-dispersed but biased, consistent with the dry bias of its long rollouts, whereas the guided DM's histogram is closer to flat with an excess at the lowest rank from its wet bias.\\
    To assess a critical characteristic of precipitation dynamics, we compute the waiting times between extreme events above the 95th percentile (Fig.~\ref{fig:precipitation_statistics}d, Fig.~\ref{fig:precipitation_stats_error}c). We find that the unconditional DM significantly underestimates the frequency of waiting times larger than 100 days. The video DM captures waiting times less than 100 days accurately, while the guided DM generates slightly more accurate waiting times that are larger than 300 days.\\
    We compute the first three empirical orthogonal functions (EOFs) using principal component analysis (PCA) up to an explained variance greater than $1\%$ and find that the guided DM is able to accurately capture the first three leading EOFs (Fig.~\ref{fig:precipitation_eofs}), while the video DM shows notable differences in the 2nd and 3rd EOF.
    We evaluate the stability of the guided and video DM with 10 simulations each over 100 years, and find that the video DM develops instabilities in terms of a shifting mean. Our guidance approach, on the other hand, is stable on centennial time scales and has a much lower global mean difference to the ERA5 ground truth (Fig.~\ref{fig:precipitation_100_years_forcing}).\\
    To further analyse the origin of the stability, we perform a perturbation experiment
    with the discriminator-guided sampling, where we scale the initial conditions before
    the rollout (Fig.~\ref{fig:restoring_force}). The two initial test frames are multiplied by 0.5 and by 2, moving their global mean from 2.8 to 1.4 and 5.6\,mm/day against a climatology of about 3\,mm/day; a third rollout starts from the end of the video DM simulation that drifted furthest in Fig.~\ref{fig:precipitation_100_years_forcing}.
    With five members over 150 days each, the global mean returns to the climatology within two steps in all cases, after which the members are indistinguishable from the unperturbed rollout.
    The diffusion-forcing model \citep{chen_diffusion_2024}, which shares its architecture, data and sampler but is trained on noised context frames, remains stable over all ten 100-year members (Fig.~\ref{fig:precipitation_100_years_forcing}), pointing to an exposure bias of the video DM as a cause for the drifting.
    We compare the spatial bias in the generated precipitation time series and find that the unconditional DM produces the smallest global mean bias (Fig.~\ref{fig:precipitation_bias}. The guidance method outperforms the video DM, the latter having the largest overall bias.\\
    The GAN baseline develops striped artefacts during the rollout (Fig.~\ref{fig:supplement_gan_precipitation_samples}) and stationary features without eastward propagation (Fig.~\ref{fig:supplement_gan_precipitation_hovmoeller}).
    Its dynamics are over-persistent (lag-one ACF 0.39 vs.\ 0.26), it overestimates waiting times beyond 300 days, and its CRPS is 11--14\% worse than that of the guided DM (Fig.~\ref{fig:supplement_gan_precipitation_stats}, \ref{fig:supplement_gan_precipitation_stats_error}).
    Its ensembles are strongly under-dispersive (spread-skill ratio 0.52--0.84, Fig.~\ref{fig:supplement_gan_precipitation_spread_skill_ratio}) and its mean absolute bias is twice that of the guided DM (Fig.~\ref{fig:supplement_gan_precipitation_bias}), although the GAN is 300 times faster to sample (Tab.~\ref{tab:compute_cost}).

\paragraph{Sample-based versus sample-free discriminators.}
    Both discriminators steer the sampler towards temporally consistent states, but they scale differently with the guidance strength (Fig.~\ref{fig:discriminator_comparison_sweeps}). For the sample-free discriminator the autocorrelation error decreases monotonically with $\lambda$ up to an optimum and degrades beyond it, reaching 0.004 at $\lambda = 20$ for vorticity and 0.005 at $\lambda = 136$ for precipitation, with the bias increasing with $\lambda$.
    Despite the large guidance scales, the guidance term remains a small correction to the prior score. Its root mean square reaches at most about 10\% of the unconditional score at the largest $\lambda$ and intermediate noise levels, and is one to two orders of magnitude smaller at the low noise levels where the fine-scale structure is formed (Fig.~\ref{fig:score_magnitudes}).
    The sample-based discriminator follows the same shape but at two to three times the error near the optimum (0.013 at its optimum for both datasets), and for precipitation its ACF error and bias vary non-monotonically along the sweep, so that no single scale optimises both metrics. 
    In the forecast ensembles at the selected scales (Fig.~\ref{fig:discriminator_comparison_vorticity}, \ref{fig:discriminator_comparison_precipitation}), the sample-free discriminator gives the lower CRPS at the first lead for both datasets (3.5\% for vorticity, 3.4\% for precipitation), where the dynamics are decided by the transition from the conditioning frames. 
    For vorticity it stays ahead through lead six before both converge to the climatological error, and its ensemble is calibrated (spread--skill ratio 0.97--1.02) whereas the sample-based one is over-dispersive at every lead (1.07). For precipitation the sample-based discriminator overtakes from lead two on by 1--2\%, with both ensembles calibrated to within 7\%. The sample-based discriminator thus matches or slightly improves the medium-range skill on precipitation, at the price of a less faithful temporal structure and the cost of generating the negative pool (Tab.~\ref{tab:compute_cost}); the sample-free discriminator captures the one-step dynamics more accurately on both datasets without any samples from the diffusion model.

\paragraph{Computational cost.}
    Tab.~\ref{tab:compute_cost} lists the training and inference cost of all models.
    Training the discriminator takes 20 GPU-hours for vorticity and 13 for
    precipitation on two H100s, against 625 and 84 GPU-hours for the unconditional
    DM. The video DM shares the architecture and training schedule of the
    unconditional DM and costs the same to train (625 and 83 GPU-hours), so
    training the full guided pipeline from scratch, unconditional DM plus
    discriminator, is 3\% and 16\% more expensive than training a video DM. The
    saving arises when an unconditional DM already exists: adding temporal
    consistency to it then costs the 20 or 13 GPU-hours of the discriminator
    instead of retraining a video DM, or the 91 GPU-hours of the diffusion-forcing
    model on precipitation. At inference the guidance adds 36\% and 89\% to the
    per-frame sampling time of the unconditional DM. The GAN samples two to three
    orders of magnitude faster than any diffusion model, which is its one advantage
    here.
    
    \begin{table}[t]
        \centering
        \caption{Computational cost of all models in GPU-hours on NVIDIA H100s (wall time times number of GPUs).
        \emph{Training}: GPU-hours up to the selected checkpoint (EMA weights with the lowest validation loss) and for the full run. Rows marked $+$ are increments added to the unconditional DM, with the dagger marking the generation cost of the sample-based discriminator.
        The guided DM row sums the unconditional DM and our discriminator and is the end-to-end cost of our method. 
        \emph{Inference}: end-to-end autoregressive rollout with all networks in single precision: seconds per frame, peak GPU memory and throughput in fields per hour at batch size one, and the maximum throughput over batch sizes with the batch size that attains it in parentheses.
        The video DM and diffusion forcing models have the unconditional DM's inference cost.}
        \label{tab:compute_cost}
        \footnotesize
        \setlength{\tabcolsep}{3pt}
        \begin{tabular}{llrrrrrr}
        \toprule
        & & \multicolumn{2}{c}{Training [GPU-h]} & \multicolumn{4}{c}{Inference, batch 1 (max.\ batch)} \\
        \cmidrule(lr){3-4} \cmidrule(lr){5-8}
        & Model & to ckpt. & full run & s/frame & GB & Fields/h & Fields/h (batch) \\
        \midrule
        \multirow{6}{*}{\rotatebox{90}{Vorticity}}
        & Video DM             & 413     & 625     & 2.2  & 0.9 & 1{,}650 & 2{,}780 (128) \\
        & Uncond.\ DM          & 155     & 625     & 2.2  & 0.9 & 1{,}650 & 2{,}780 (128) \\
        & $+$ Disc.\ (ours)    & $+$6.4  & $+$20   & $+$0.9 & $+$0.03 & --   & -- \\
        & $=$ Guided DM        & 161     & 645     & 3.1  & 1.0 & 1{,}180 & 2{,}450 (64) \\
        & $+$ Disc.\ (sample-based) & $+$23$^\dagger$ $+$ 12 & $+$23$^\dagger$ $+$ 21 & $+$0.9 & $+$0.03 & -- & -- \\
        & GAN                  & 37      & 74      & 0.03 & 1.0 & 119{,}000 & 270{,}000 (64) \\
        \midrule
        \multirow{7}{*}{\rotatebox{90}{Precipitation}}
        & Video DM             & 13      & 83      & 3.5  & 0.9 & 1{,}020 & 1{,}700 (64) \\
        & Diffusion forcing    & 37      & 91      & 3.5  & 0.9 & 1{,}020 & 1{,}700 (64) \\
        & Uncond.\ DM          & 22      & 84      & 3.5  & 0.9 & 1{,}020 & 1{,}700 (64) \\
        & $+$ Disc.\ (ours)    & $+$3.7  & $+$13   & $+$3.1 & $+$0.8 & --   & -- \\
        & $=$ Guided DM        & 26      & 97      & 6.6  & 1.7 & 550     & 840 (32) \\
        & $+$ Disc.\ (sample-based) & $+$19$^\dagger$ $+$ 6.2 & $+$19$^\dagger$ $+$ 10 & $+$3.1 & $+$0.8 & -- & -- \\
        & GAN                  & 15      & 36      & 0.02 & 0.4 & 186{,}000 & 695{,}000 (128) \\
        \bottomrule
        \end{tabular}
    \end{table}

\section{Discussion}
    \label{sec:discussion}
    We propose a time consistency discriminator that guides the sampling process of unconditionally trained image diffusion models (DMs) to generate time-consistent image sequences, i.e., dynamically realistic simulations and videos. Our discriminator is trained separately from the diffusion model on the target data only and hence independent of the DM architecture.\\ 
    We evaluate our method on two challenging datasets with complex non-linear dynamics, namely 2D Navier-Stokes turbulence simulations and global precipitation reanalysis.
    We find that the discriminator guidance enables the unconditionally trained DM to generate realistic dynamics with comparable skill to a video DM trained from scratch. While the video DM produces more accurate short-term forecasts, the guidance method matches it on longer lead times and has lower biases and improved stability. 
    Our method enables stable rollouts over 100 years and is not subject to unstable mean shifts as the video DM, promising immense potential for climate research applications \citep{chattopadhyay_challenges_2024, wang_accurate_2018}.
    This likely follows from the construction of the sampler: each frame is drawn from the unconditional DM, which represents the climatology, and the discriminator only selects states consistent with the previous frames. 
    The guidance thus acts as a restoring force towards the marginal at every step and cannot accumulate a drift, whereas the video DM conditions on its own
    imperfect outputs with no such anchor. 
    This is in line with our finding that diffusion-forcing training
    \citep{chen_diffusion_2024} removes the drift of the video DM: by noising the context
    frames during training, the model learns to condition on imperfect inputs, which is
    exactly the situation it faces in an autoregressive rollout.
    Our method is computationally efficient (see Appendix \ref{app:discriminator_training}) and increases the reusability of pre-trained image diffusion models to a wider range of tasks that require temporal consistency without the need for costly retraining video models from scratch, making generative modeling more sustainable and accessible.
    Moreover, our guidance method is promising for long rollouts of video DMs, which we leave for future research.\\
    We only consider univariate simulations here, but, we believe that extensions to multiple variables are straightforward. While our method produces more realistic long-term simulations, it has a lower short-term forecast skill than the video DM. Further explorations in terms of the discriminator architecture and training might enable improvements in that respect.
    In this work we train the diffusion image models from scratch and do not consider existing off-the-shelf open weight models. Such an extension will be an important next step, extending also to more general settings of natural images.
    We apply our discriminator guidance method only to one type of diffusion model and sampler \citep{karras_elucidating_2022}, but it should, in principle, be applicable to others as well \citep{kim_refining_2023}.
    This study focuses on video \emph{generation}, however, the guidance method is also applicable to enforce time-consistency in video \emph{processing} tasks such as super-resolution or inpainting, which is crucial in related applications in weather and climate \citep{plesiat_artificial_2024, mardani_residual_2025, addison_machine_2026, hess_fast_2025, aich_conditional_2026}. We hope that our results will encourage further research on video synthesis with discriminator-guided diffusion models.

\acks{
The authors would like to thank Sebastian Bathiany, Alistair White and Maha Badri for their helpful comments and discussions of the work.
NB and MG acknowledge funding by the Volkswagen Foundation.
MA acknowledges funding under the Excellence Strategy of the Federal Government and the L\"ander through the TUM Innovation Network EarthCare. 
This is ClimTip contribution \#220; the ClimTip project has received funding from the European Union's Horizon Europe research and innovation programme under grant agreement No. 101137601.
PH is funded by the Deutsche Forschungsgemeinschaft (DFG, German Research Foundation) project number 556377191.
YH acknowledges funding by the Alexander von Humboldt Fundation. 
The authors acknowledge the European Regional Development Fund (ERDF), the German Federal Ministry of Education and Research, and the Land Brandenburg for supporting this project by providing resources on the high-performance computer system at the Potsdam Institute for Climate Impact Research.
}

\appendix
\section{Diffusion models}
\label{app:diffusion_model}
    For all diffusion models trained in this work, we use the DDPM++ UNet \citep{song_scorebased_2021} with hyperparameters adapted as shown in Tab.~\ref{tab:diffusion_config}. The UNets have around 33.8M parameters for the vorticity configuration and 25.7M parameters in the precipitation configuration. For inference, we use weights from an exponential moving average (EMA). We extend the architecture by applying periodic padding in both spatial dimensions for the vorticity simulation and in the longitude direction for the ERA5 precipitation data. We use the EDM preconditioning from Karras et al. 2022~\citep{karras_elucidating_2022}, which is given for the unconditional DM by 
    \begin{equation}
    \bm{S}_\phi(\bm{x}_t; t) := c_{\text{skip}}(t)\bm{x}_t + c_{\text{out}}(t) \bm{f}_{\phi}(c_{\text{in}}(t)\bm{x}_t; c_{\text{noise}}(t)),
    \end{equation}
    where $\bm{f}_\phi(\cdot)$ is the UNet denoiser network and the coefficients are defined in \citep{karras_elucidating_2022}.
    For training, we use a log-normal distribution to sample the noise levels $\ln(\sigma) \sim \mathcal{N}(P_{\text{mean}}, P_{\text{std}}^2)$ \citep{karras_elucidating_2022}.
    For sampling, we also use the EDM noise schedule \citep{karras_elucidating_2022}, defined as 
    \begin{equation}
    \sigma_i := \left( \sigma_{\text{max}}^{1/\rho} + \frac{i}{N-1} \left(\sigma_{\text{min}}^{1/\rho} - \sigma_{\text{max}}^{1/\rho}\right)  \right)^\rho,
    \label{eq:noise_schedule}
    \end{equation}
    where $i \in \{ 0,...,N-1\}$, $N$ is the number of sampling steps and $\sigma_{\text{min}}$, $\sigma_{\text{max}}$, $\rho$ are defined in Tab.~\ref{tab:diffusion_config}.
    We use the stochastic EDM sampler \citep{karras_elucidating_2022} with the parameters given in Tab.~\ref{tab:diffusion_config} for the video, unconditional and guided diffusion model. 
\begin{table}
 \caption{Configuration details of the diffusion model architectures, training and sampling.}
  \centering
  \begin{tabular}{lll}
    \toprule
                            & 2D vorticity & ERA5 precipitation \\
    \midrule
    \textbf{Architecture}   &               &  \\
    Input dimension (unconditional)        & (B,1,256,256) &  (B,1,180,360) \\
    Input dimension  (video)  & (B,3,256,256) &  (B,3,180,360) \\
    Output dimension        & (B,1,256,256)         &  (B,1,180,360) \\
    Num. Resnet blocks      &  3            &  2 \\
    Num. attention blocks   &  3            &  2 \\
    Attention resolution    &  8,4          &  8,4 \\
    Channel                 &  (1,2,2)      & (1,2,2) \\
    Channel multiplier      &  128          & 128 \\
    \midrule
    \textbf{Training}       &               &  \\
    Batch size (B)          & 2             & 2 \\
    Learning rate (LR)      & $10^{-4}$     & $10^{-4}$\\
    Optimizer               & AdamW         & AdamW \\
    Epochs                  & 350           & 250 \\
    EMA rate                & 0.9999        & 0.9999\\
    \midrule
    \textbf{Sampling}       &               &  \\
    $\sigma_{\text{min}}$   & 0.002         & 0.002 \\
    $\sigma_{\text{max}}$   & 80            & 80 \\
    $\sigma_{\text{data}}$  & 0.5           & 0.5 \\
    $\rho$                  & 7             & 7 \\
    $P_{\text{mean}}$       & -1.2          & -1.2 \\
    $P_{\text{std}}$        & 1.2           & 1.2 \\
    $S_{\text{t}_{\text{min}}}$  & 0        & 0 \\
    $S_{\text{t}_{\text{max}}}$  & 1000     & 1000 \\
    $S_{\text{noise}}$      & 1.005         & 1.0 \\
    $S_{\text{churn}}$      & 55            & 80 \\
    Num. steps              & 50            & 100 \\
    $\lambda$ (guidance strength)    & 20      & 68 \\
    \bottomrule
  \end{tabular}
  \label{tab:diffusion_config}
\end{table}

\subsection{Diffusion forcing.}
\label{app:diffusion_forcing}
    The diffusion-forcing baseline \citep{chen_diffusion_2024} is the video DM retrained with noised context. 
    The architecture, data, optimiser and sampler are those of the video DM (Tab.~\ref{tab:diffusion_config}), the only change is the training
    objective. 
    For each training sample, every context frame independently receives
    diffusion noise at a level drawn from the EDM log-normal distribution, or is left
    clean with probability 0.25 so that the clean-history rollout stays inside the
    training distribution, and the context noise levels are passed to the network as
    additional noise-level embeddings. 
    At inference the model is rolled out autoregressively on its own clean outputs, as the video DM. Training took 91 GPU-hours for 650 epochs, with the selected checkpoint at epoch 261, compared to 83 GPU-hours for 250 epochs of the video DM (Tab.~\ref{tab:compute_cost}).

\section{Discriminator model}

\subsection{Time-consistency guidance}
\label{app:derivaion}

    Given a time series of length $N$, $\{\bm{x}^n | n=1,2,...,N\}$, containing images $\bm{x}^n \in \mathbb{R}^{B \times C \times H \times W}$ that are temporally ordered, we aim to train a discriminator network to classify whether a shown sample $\bm{x}^{n+1}$ is in temporal order with respect to the current and a sequence of $m$ previous time frames $\{\bm{x}^{n-m},\bm{x}^{n-m+1}, ..., \bm{x}^{n}\} = \bm{x}^{(n-m):n}$ or not.
    In other words, whether a sample is drawn from the conditional distribution $\bm{x}^{n+1} \sim p(\bm{x}^{n+1}|\bm{x}^{(n-m):n})$ or the unconditional distribution $\bm{x}^{n+1} \sim p(\bm{x}^{n+1})$. Here and in the following we drop the explicit dependency on the noise time $t$. 
    As shown in \citep{goodfellow_generative_2014}, an optimal discriminator with parameters $\theta$ can then be written as
    \begin{equation}
    D_\theta(\bm{x}^{n+1}; \bm{x}^{(n-m):n}) = \frac{p(\bm{x}^{n+1}|\bm{x}^{(n-m):n})}{p(\bm{x}^{n+1}|\bm{x}^{(n-m):n}) + p(\bm{x}^{n+1})},
    \end{equation}
    which we can rewrite as
    \begin{equation}
	    \frac{D_{\theta}(\bm{x}^{n+1}; \bm{x}^{(n-m):n})}{1-D_{\theta}(\bm{x}^{n+1};\bm{x}^{(n-m):n})} = \frac{p(\bm{x}^{n+1}|\bm{x}^{(n-m):n})}{p(\bm{x}^{n+1})}.
    \end{equation}
    Taking the log and computing the gradient with respect to $\bm{x}^{n+1}$ gives
    \begin{equation}
    \nabla_{\bm{x}^{n+1}} \log \left(\frac{D_\theta(\bm{x}^{n+1}; \bm{x}^{(n-m):n})}{1-D_\theta(\bm{x}^{n+1}; \bm{x}^{(n-m):n})} \right) =  \nabla_{\bm{x}^{n+1}} \log \left(\frac{p(\bm{x}^{n+1}|\bm{x}^{(n-m):n})}{p(\bm{x}^{n+1})} \right),
    \end{equation}
    which we use for our guidance term:
    \begin{equation}
   \bm{d}_\theta(\bm{x}^{n+1}; \bm{x}^{(n-m):n}) :=  \nabla_{\bm{x}^{n+1}} \log \left(\frac{p(\bm{x}^{n+1}|\bm{x}^{(n-m):n})}{p(\bm{x}^{n+1})} \right) .
    \label{eq:discriminator_guidance_v2}
    \end{equation}
    Using the expression
    \begin{equation}
        p(\bm{x}^{n+1}|\bm{x}^{(n-m):n}) = p(\bm{x}^{n+1})\frac{p(\bm{x}^{n+1}|\bm{x}^{(n-m):n})}{p(\bm{x}^{n+1})},
    \end{equation}
    and computing the score functions gives 
    \begin{align}
    \begin{split}
    \nabla_{\bm{x}^{n+1}} \log p(\bm{x}^{n+1}|\bm{x}^{(n-m):n}) & = \nabla_{\bm{x}^{n+1}} \log p(\bm{x}^{n+1}) \\
    &+ \nabla_{\bm{x}^{n+1}} \log \left(\frac{p(\bm{x}^{n+1}|\bm{x}^{(n-m):n})}{p(\bm{x}^{n+1})} \right).
    \end{split}
    \label{eq:conditional_score}
    \end{align}
    From Eq.~\ref{eq:conditional_score} we see that we can approximate the conditional score $\nabla_{\bm{x}^{n+1}} \log p(\bm{x}^{n+1}|\bm{x}^{(n-m):n})$ with an unconditional score model $\nabla_{\bm{x}^{n+1}} \log p(\bm{x}^{n+1}) \approx \bm{s}_\theta(\bm{x}^{n+1})$ and the guidance in Eq.~\ref{eq:discriminator_guidance_v2}, as
    \begin{equation}
    \nabla_{\bm{x}^{n+1}} \log p(\bm{x}^{n+1}|\bm{x}^{(n-m):n})  \approx 
     \bm{s}_\theta(\bm{x}^{n+1}) + \bm{d}_\theta(\bm{x}^{n+1}; \bm{x}^{(n-m):n}).
    \end{equation}

\subsection{Network and training}
\label{app:discriminator_training}

    We adapt the noise time-conditioned encoder part of the UNet from \citep{kim_refining_2023} for the discriminator model with a two-layer fully connected network as a decoder (see Tab.~\ref{tab:discriminator_config} for hyperparameter configurations). We add periodic padding in both spatial dimensions for the vorticity simulation and in the longitude direction for the ERA5 precipitation data.
    The discriminator is trained on random crops of the fields rather than on the full domain. For each training batch one crop window is drawn and applied identically to the candidate frame and to its context frames, so that the temporal relation inside the window is preserved. The window height and width are sampled independently from
    $\{32, 64, 128\}$ and $\{32, 64, 128, 256\}$ grid cells for precipitation and vorticity, and its position is
    uniform over the field. At inference the discriminator is evaluated on the full field.
    The discriminator network has 1.15M parameters in the vorticity configuration and 7.87M in the precipitation configuration, making it around 29.4 and 3.3 times smaller, respectively, than the diffusion networks.

\subsection{Guided sampling}    
\label{app:sampler}
    We adapt the stochastic discriminator guidance sampler \citep{karras_elucidating_2022, kim_refining_2023}, which solves the sampling ODE with stochastic churn with a second-order accurate solver. Stochasticity is controlled through the parameters $S_{\text{noise}}, S_{\text{churn}}, S_{\text{t}_{\text{max}}}, S_{\text{t}_{\text{max}}}$ in the coefficient $\gamma_i$ with
    \begin{equation}
    \gamma_i = \begin{cases}
                    \min \left( \frac{S_{\textrm{churn}}}{T}, \sqrt{2}-1 \right) & \text{if } t_i \in [S_{\text{t}_{\text{min}}},S_{\text{t}_{\text{max}}} ], \\
                    0 & \text{otherwise}.
            \end{cases}
    \end{equation}
    We provide pseudocode of the implementation in Alg.~\ref{alg:sampler}.
    \begin{algorithm}
        \caption{Time-consistency (TC) guided sampling (adapted from \citep{karras_elucidating_2022, kim_refining_2023})}
        \begin{algorithmic}[1]
        \State \textbf{input:} $\bm{S}_\phi$, $D_\theta$, $\bm{x}^{n}$, $\bm{x}^{n-1}$, $\lambda$, $t_{i\in\{ 0,...,T\}}$, $\gamma_{i\in\{ 0,...,T-1\}}$, $S_{\textrm{noise}}$.
        \State \textbf{sample} $\mathbf{x}^{n+1}_T \sim \mathcal{N}(\mathbf{0}, t^2_T\mathbf{I})$
        \For{$i = 0$ to $T$}
            \State \textbf{sample} $\bm{\epsilon}_i \sim \mathcal{N}\left(\mathbf{0}, S^2_{\mathrm{noise}} \mathbf{I}\right)$
            \State $\hat{t}_i \leftarrow t_i + \gamma_i t_i$
            \State $\hat{\bm{x}}^{n+1}_i \leftarrow \bm{x}^{n+1}_i + \sqrt{\hat{t}^2_i - t_i^2} \bm{\epsilon}_i$
            \State $\bm{s}_i \leftarrow (\hat{\bm{x}}^{n+1}_i - \bm{S}_\phi(\hat{\bm{x}}^{n+1}_i; \hat{t}_i)/\hat{t}_i$
            \State $\bm{d}_i \leftarrow - t_i \nabla_{\hat{\bm{x}}^{n+1}_i} \log \left(\frac{\bm{D}_\theta(\hat{\bm{x}}^{n+1}_i; \bm{x}^{n}, \bm{x}^{n-1}, \hat{t}_i)}{1-\bm{D}_\theta(\hat{\bm{x}}^{n+1}_i; \bm{x}^{n}, \bm{x}^{n-1}, \hat{t}_i)} \right)$
            \Comment{TC guidance}
            \State $\bm{x}_{i+1} \leftarrow \hat{\bm{x}}_i + (t_{i+1} - \hat{t}_i)(\bm{s_i} + \lambda \bm{d}_i)$
            \If{$t_{i+1} \neq 0$} 
            \State $\bm{s}'_i \leftarrow (\bm{x}^{n+1}_{i+1} - \bm{S}_\phi(\bm{x}^{n+1}_{i+1}; \hat{t}_{i+1}))/\hat{t}_{i+1}$
            \State $\bm{d}'_i \leftarrow - t_{i+1} \nabla_{\bm{x}^{n+1}_{i+1}} \log \left(\frac{\bm{D}_\theta(\bm{x}^{n+1}_{i+1}; \bm{x}^{n}, \bm{x}^{n-1}, t_{i+1})}{1-\bm{D}_\theta(\bm{x}_{i+1}^{n+1}; \bm{x}^{n}, \bm{x}^{n-1}, t_{i+1})} \right)$
            \Comment{TC guidance}
            \State $\bm{x}_{i+1}^{n+1} \leftarrow \hat{\bm{x}}_i^{n+1} + (t_{i+1} - \hat{t}_i) \left[(\frac{1}{2}\bm{s_i}+ \lambda \bm{d}_i) + \frac{1}{2}(\bm{s}'_i + \lambda \bm{d}'_i) \right]$
            \EndIf
        \EndFor
        \State \textbf{return} $\mathbf{x}^{n+1}_T$
        \end{algorithmic}
        \label{alg:sampler}
    \end{algorithm}

\begin{table}
 \caption{Discriminator model architecture and training parameters.}
  \centering
  \begin{tabular}{lll}
    \toprule
                            & 2D vorticity & ERA5 precipitation \\
    \midrule
    \textbf{Architecture}   &               &  \\
    Input dimension         & (B,3,256,256) & (B,3,180,360) \\
    Output dimension        & (B,1)         & (B,1) \\
    Num. Resnet blocks      & 2            & 2 \\
    Num. attention blocks   & 2            & 2 \\
    Attention resolution    & 8,4          & 8,4 \\
    Channel                 & (1,2,2)      & (4,2,1) \\
    Channel multiplier      & 128          & 64 \\
    MLP layer size          & 1024         & 1024 \\
    Num. MLP layer          & 2            & 2 \\
    \midrule
    \textbf{Training}       &               &  \\
    Batch size (B)          & 8             & 8 \\
    Learning rate (LR)      & $10^{-4}$     & $10^{-4}$\\
    Optimizer               & AdamW         & AdamW \\
    Epochs                  & 500           & 500 \\
    \bottomrule
  \end{tabular}
  \label{tab:discriminator_config}
\end{table}

\subsection{Generative adversarial network}
    \label{app:gan}
    A conditional GAN \cite{goodfellow_generative_2014} is included as a baseline following  the design of DGMR \cite{ravuri_skilful_2021a} and DVD-GAN \cite{clark_adversarial_2019}.
    The generator $G(\bm{x}^{n}_0, \bm{x}^{n-1}_0, \bm{z})$ is the U-Net of the video DM 
    at identical capacity for vorticity (34.0M against 33.8M parameters) and at half the channel width for precipitation (64 instead of 128 base channels, 6.5M against 25.7M parameters), with a spatially coherent latent $\bm{z}$ drawn at one eighth of the grid resolution and refined by three convolutions in place of the noised input frame. 
    A temporal discriminator, the adversarial counterpart of the time-consistency discriminator with the same encoder, judges the candidate frame together with its two past frames. Both return a logit per patch and use spectral normalisation.
    Training follows \cite{ravuri_skilful_2021a}: hinge losses, a grid-cell regulariser on the mean of two latent draws, two-timescale Adam, a lazy R1 penalty, an initial discriminator-only phase, and an exponential moving average of the generator for sampling.
    The vorticity configuration additionally needs a global latent vector, a 128-dimensional latent vector injected into the U-Net embedding at every resolution, in addition to the spatial latent field, and a down-weighted, widened spatial discriminator, without which the generator collapses to a deterministic map within the first epoch (Tab.~\ref{tab:gan_config}).
    Checkpoints are chosen whose 600-step autoregressive rollout stays bounded as the validation CRPS, scored at short lead times, does not detect this divergence.
    \begin{table}[h]
    \centering
    \caption{GAN baseline configurations. The generator backbone is the video DM U-Net
    of the respective dataset. Discriminators are the encoder of the time-consistency
    discriminator with a patch output. Channel multipliers and residual blocks per
    level were chosen as for the U-Nets of Tab.~\ref{tab:diffusion_config}.}
    \label{tab:gan_config}
    \begin{tabular}{lcc}
        \toprule
                                            & Precipitation     & Vorticity \\
        \midrule
        Generator parameters                & 6.5M              & 34.0M \\
        Spatial latent (channels, downsampling) & 8, $\times 8$  & 8, $\times 8$ \\
        Global latent vector                & --                & 128 \\
        Temporal discriminator (channels, mult., blocks) & 64, [4,2,1], 2 & 128, [1,2,2], 3 \\
        Spatial discriminator (channels, mult., blocks)  & 32, [4,2,1], 2 & 128, [1,2,2], 3 \\
        discriminator parameters (temporal, spatial) & 4.6M, 1.2M     & 12.7M, 12.7M \\
        Spatial discriminator weight               & 1.0               & 0.25 \\
        Discriminator input                        & 64-row latitude band & full field \\
        Batch size                          & 8                 & 2 \\
        $K$, $\lambda_{\text{reg}}$         & 2, 2              & 2, 2 \\
        Selected checkpoint                 & epoch 60, 60k updates & epoch 3, 88k updates \\
        \bottomrule
    \end{tabular}
\end{table}

\section{Data}

\subsection{2D Navier-Stokes experiments}
\label{app:navier_stokes_data}
    We perform numerical simulations of the two-dimensional incompressible Navier-Stokes equations in vorticity stream function formulation using the GeophysicalFlow.jl Julia package \citep{constantinou_geophysicalflowsjl_2021}. The simulation uses periodic boundary conditions, hyperviscosity $\nu=2e^{-7}$ of second order, a linear drag coefficient of $\mu=1e^{-1}$ and integration time step $\Delta t=0.005$. We subsample the simulation saving every fourth time step to disk. We apply stochastic forcing defined with an Ornstein-Uhlenbeck process and a forcing wavenumber $k_f = 6 \cdot2\pi/L$, where $L=256$, forcing bandwidth $\delta_f=1.5 \cdot 2\pi /L$ and an energy input rate of $\epsilon = 0.1$.  We wait for 500 steps for the simulation to reach a statistical equilibrium. We then standardize the data by subtracting the mean and dividing by the standard deviation for training the diffusion and discriminator networks.

\subsection{Precipitation data}
\label{app:precipitation_data}
    As a challenging, real-world application, we use global daily precipitation fields from the ERA5 reanalysis dataset \citep{hersbach_era5_2020}. The data is openly available for download 
    at the Copernicus Climate Data Store (\url{https://cds.climate.copernicus.eu/datasets/reanalysis-era5-single-levels?tab=overview}). We regrid the data to $1^\circ$ horizontal spatial resolution using bilinear interpolation. As additional preprocessing steps we apply a log-transform with $\tilde{x} = \log(x + \epsilon) - \log(\epsilon)$, $\epsilon = 10^{-4}$, and normalization approximately into the range $[-1,1]$ for the diffusion and discriminator networks.
    We split the daily data into periods of 1979-2000, 2001-2010 and 2011-2020 for training, validation and testing.

\section{Evaluation metrics}
    \label{app:metrics}
    We denote the ground truth and predicted spatial fields with $y_{k,l}^n$ and $x_{k,l}^n$, respectively, where $n=1,...,N$ is the time index, $k=1,...,K$ is the height (or latitude) index and $l=1,...,L$ is the width (or longitude) index.
    We weigh the spherical data with a factor
    $$w(k) = \frac{\cos(\text{lat}(k))}{\frac{1}{K} \sum_{i=1}^K\cos(\text{lat}(i))}$$
    that accounts for the spherical geometry of the precipitation data and set $w(k) = 1$ for the vorticity experiments.

\subsection{Deterministic metrics}

    \paragraph{Root mean square error.} We define a spatially weighted root mean square error (RMSE) as
    \begin{equation}
    	\text{RMSE} := \sqrt{\frac{1}{L} \sum_{l=1}^L \frac{1}{K} \sum_{k=1}^K w(k) \sum_{n=1}^N (y^n_{k,l} -  x^n_{k,l})^2}.
    \label{eq:rmse}
    \end{equation}
    
    \paragraph{Bias.} The bias at each spatial location is defined as
    \begin{equation}
    	\text{Bias}_{k,l} := \frac{1}{N} \sum_{n=1}^N (y^n_{k,l} -  x^n_{k,l}).
        \label{eq:bias}
    \end{equation}
    
    \paragraph{Autocorrelation function.} The global mean over local autocorrelation functions (ACFs) is calculated by first removing the monthly mean for seasonal adjustment and standardizing the time series and then computing the ACF with
    \begin{equation}
    	\text{ACF}(j) := \frac{1}{L} \sum_{l=1}^L \frac{1}{K} \sum_{k=1}^K w(k)\frac{\frac{1}{N}\sum_{n=1}^N[(x^n_{k,l} - \bar{x}_{k,l})(x^{n-j}_{k,l} - \bar{x}_{k,l})]}{\sigma^2_{k,l}},
    \label{eq:autocorrelation}
    \end{equation}
    where the bar denotes the temporal mean and $\sigma^2_{k,l}$ is the variance at a spatial location.
    
    \paragraph{Wasserstein distance.} We use the Wasserstein distances to compute changes between consecutive rows in the Hovmöller diagrams, which allow us to quantitatively asses their similarity. 
    We, therefore, treat two rows with consecutive time steps as two tuples of probabilities, $(p_1,..., p_K)$ and $(q_1,..., q_K)$, by taking their absolute value and normalizing them to sum to 1. We then compute the Wasserstein-1 distance with 
    \begin{equation}
    	\text{W}_1(P,Q) :=  \frac{1}{N} \sum_{i=1}^N |F_p(i) - F_q(i)|,
    \label{eq:wasserstein_distance}
    \end{equation}
    where $F_p$ and $F_q$ denote the cumulative distribution functions.

\subsection{Probabilistic metrics}

    \paragraph{Continuous ranked probability score.} The continuous ranked probability score (CRPS) is computed for an ensemble size $B$ at a given time step $n$ following \citep{price_probabilistic_2025},
    
    \begin{equation}
    	\text{CRPS}^n := \frac{1}{L} \sum_{l=1}^L \frac{1}{K} \sum_{k=1}^K w(k) \left( \frac{1}{B} \sum_{b=1}^B \left|x^{n,b}_{k,l} - y^{n,b}_{k,l} \right| - \frac{1}{2B^2} \sum_{b=1}^B \sum_{b'=1}^B \left|x^{n,b}_{k,l} - x^{n,b'}_{k,l} \right| \right),
    \label{eq:crps}
    \end{equation}
    and a lower CRPS score is better.
    
    \paragraph{Spread skill ratio.} The ensemble spread for a single time step $n$ is defined, as in \citep{price_probabilistic_2025},
    
    \begin{equation}
    	\text{Spread}^n := \sqrt{\frac{1}{L} \sum_{l=1}^L \frac{1}{K} \sum_{k=1}^K w(k) \frac{1}{B-1} \sum_{b=1}^B \left( x^{n,b}_{k,l} - \tilde{x}^n_{k,l}\right)^2},
    \label{eq:spread}
    \end{equation}
    where $\tilde{x}^n_{k,l}$ is the ensemble mean. The ensemble skill at time step $n$ is then defined as
    \begin{equation}
    	\text{Skill}^n := \sqrt{\frac{1}{L} \sum_{l=1}^L \frac{1}{K} \sum_{k=1}^K w(k) \left( y^n_{k,l} - \tilde{x}^n_{k,l}\right)^2}.
    \label{eq:skill}
    \end{equation}
    Assuming that the ensemble members are all exchangeable, the spread skill ratio is then defined \citep{fortin_why_2014, price_probabilistic_2025},
    \begin{equation}
    	\text{Spread-skill-ratio} := \sqrt{\frac{M+1}{M}} \frac{\text{Spread}}{\text{Skill}},
    \label{eq:spread_skill_ratio}
    \end{equation}
    which should be close to 1 for a perfect forecast.
    
    
\section{Additional results}

\subsection{Samples, calibration and error diagnostics}

\begin{figure}[H]
    \centering
    \includegraphics[width=1.0\linewidth]{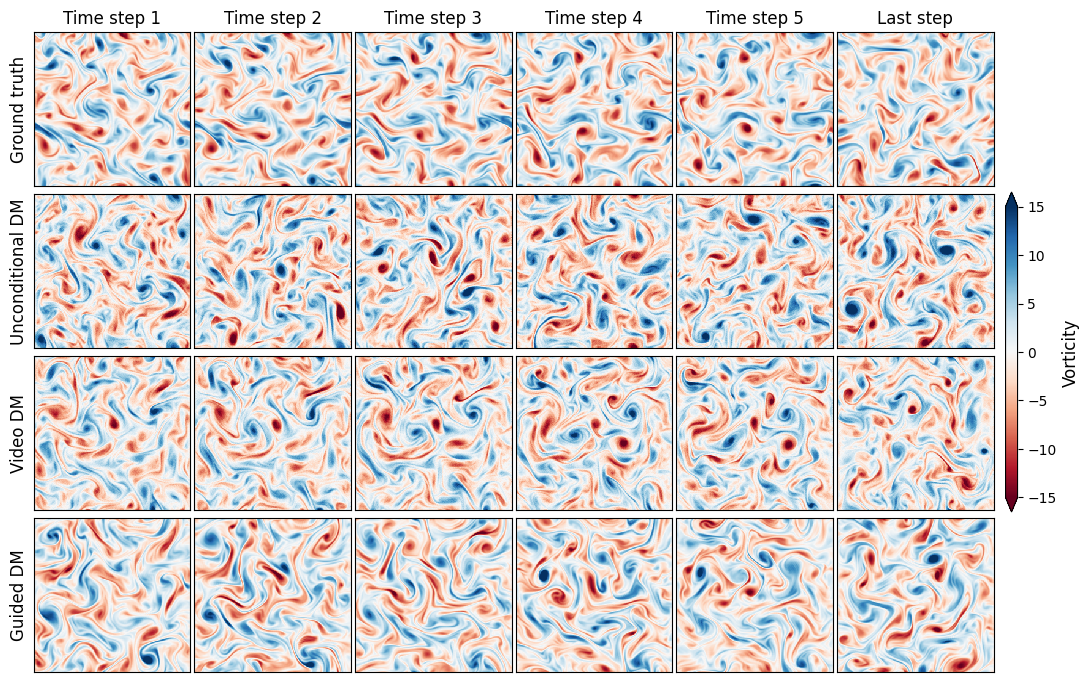}
    \caption{\textbf{Qualitative comparison} of the first five and last 2D vorticity fields from the direct numerical Navier-Stokes turbulence simulation (top), unconditional DM (upper middle), video DM (lower middle) and our discriminator guidance DM (bottom). Each row shows a single rollout starting from the same initial condition. Note that the pairing between the generated and ground truth samples decreases due to the chaotic dynamics.}
    \label{fig:voriticity_samples}
\end{figure}

\begin{figure}[H]
    \centering
    \includegraphics[width=1.0\linewidth]{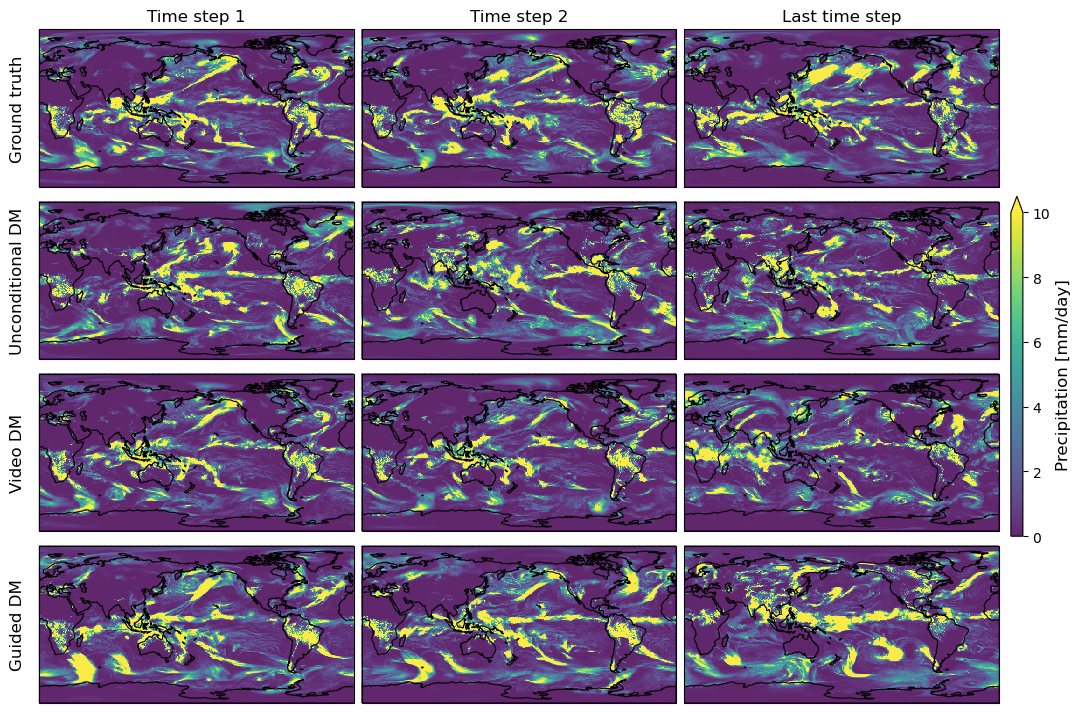}
    \caption{\textbf{Qualitative comparison} of the first two and last daily precipitation fields from the ERA5 ground truth (top), unconditional DM (upper middle), video DM (lower middle) and our discriminator guidance DM (bottom).}
    \label{fig:precipitation_samples}
\end{figure}

\begin{figure}[H]
    \centering
    \includegraphics[width=0.65\linewidth]{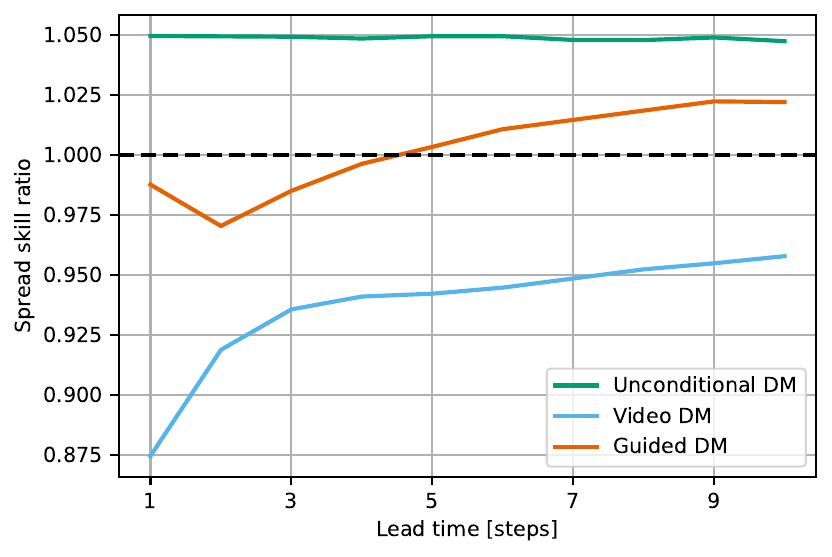}
    \caption{\textbf{The spread skill ratio} (SSR) of vorticity forecasts is shown for 100 ensemble forecasts with 50 members and 10-step lead time for the (green) unconditional DM, (blue) video DM and (red) guided DM. A perfect forecast would have a SSR of one.}
    \label{fig:vorticity_ssr}
\end{figure}

\begin{figure}[H]
    \centering
    \includegraphics[width=0.9\linewidth]{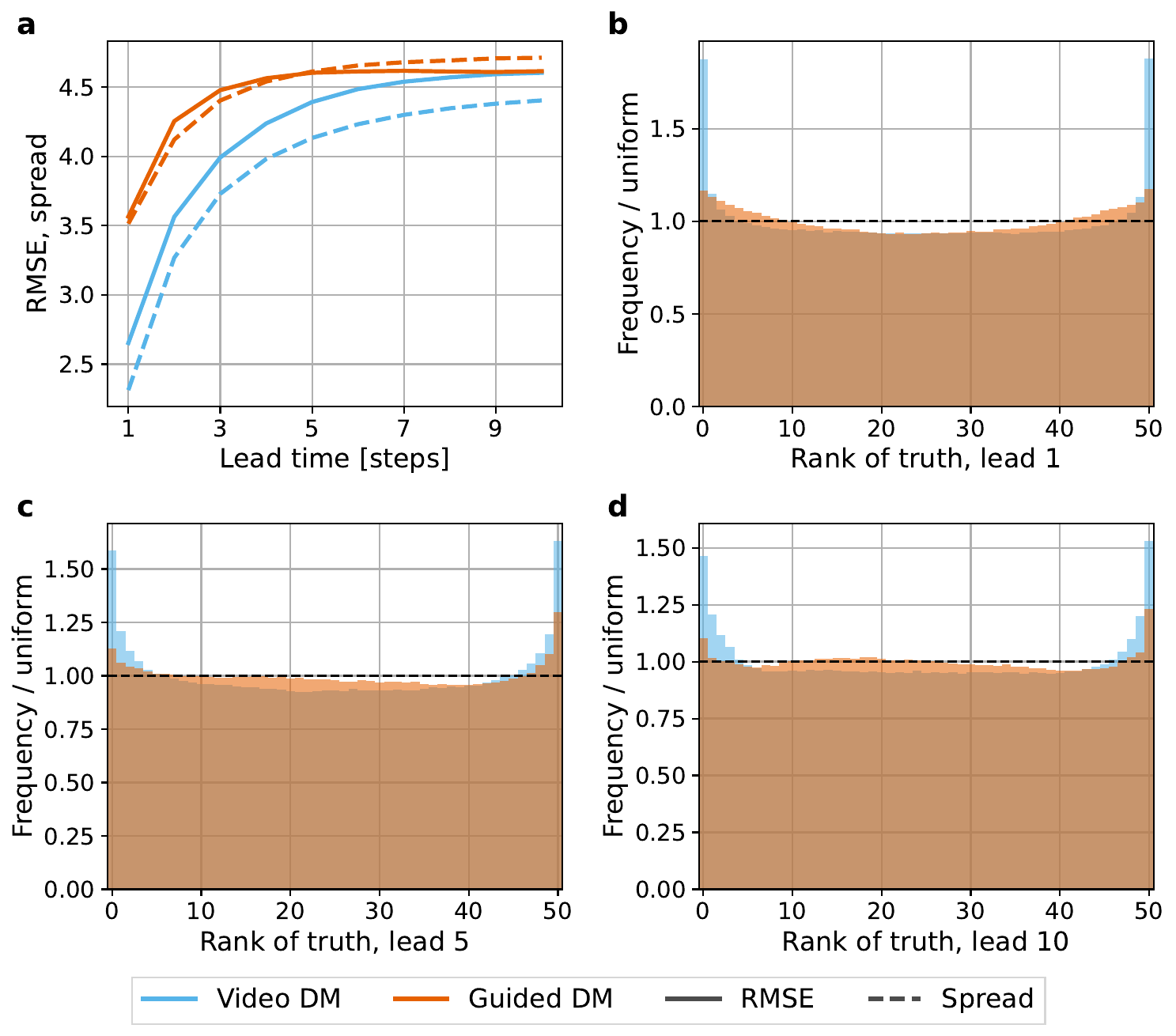}
    \caption{\textbf{Uncertainty calibration} of vorticity forecasts is shown for 100 ensemble forecasts with 50 members and 10-step lead time for the video DM (blue) and the guided DM (red). The RMSE (solid) and ensemble spread (dashed) are shown as a function of lead time (a), and rank histograms of the ground truth within the ensemble, normalized by the uniform distribution (black dashed), at lead times one (b), five (c) and ten (d).}
    \label{fig:calibration_vorticity}
\end{figure}

\begin{figure}[H]
    \centering
    \includegraphics[width=0.65\linewidth]{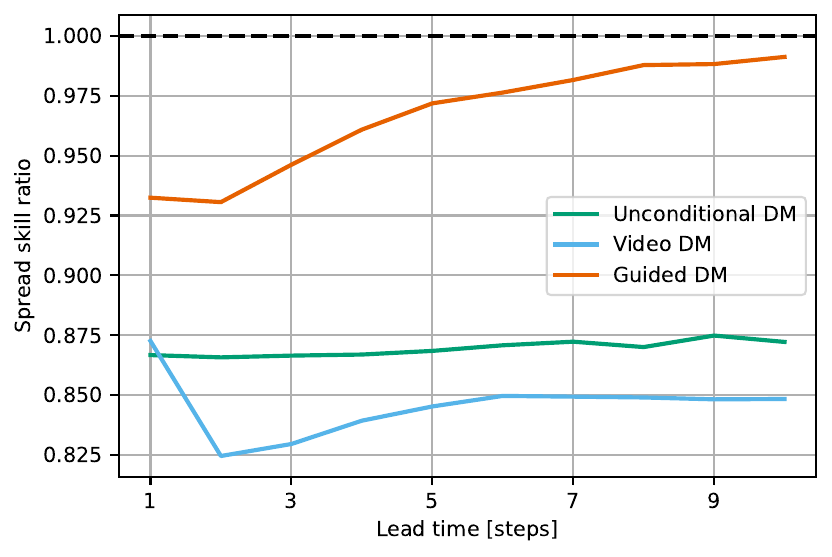}
    \caption{\textbf{The spread skill ratio} (SSR) of precipitation forecasts is shown for 100 ensemble forecasts with 50 members and 10-step lead time for the (green) unconditional DM, (blue) video DM and (red) guided DM.}
    \label{fig:precipitation_ssr}
\end{figure}

\begin{figure}[H]
    \centering
    \includegraphics[width=0.9\linewidth]{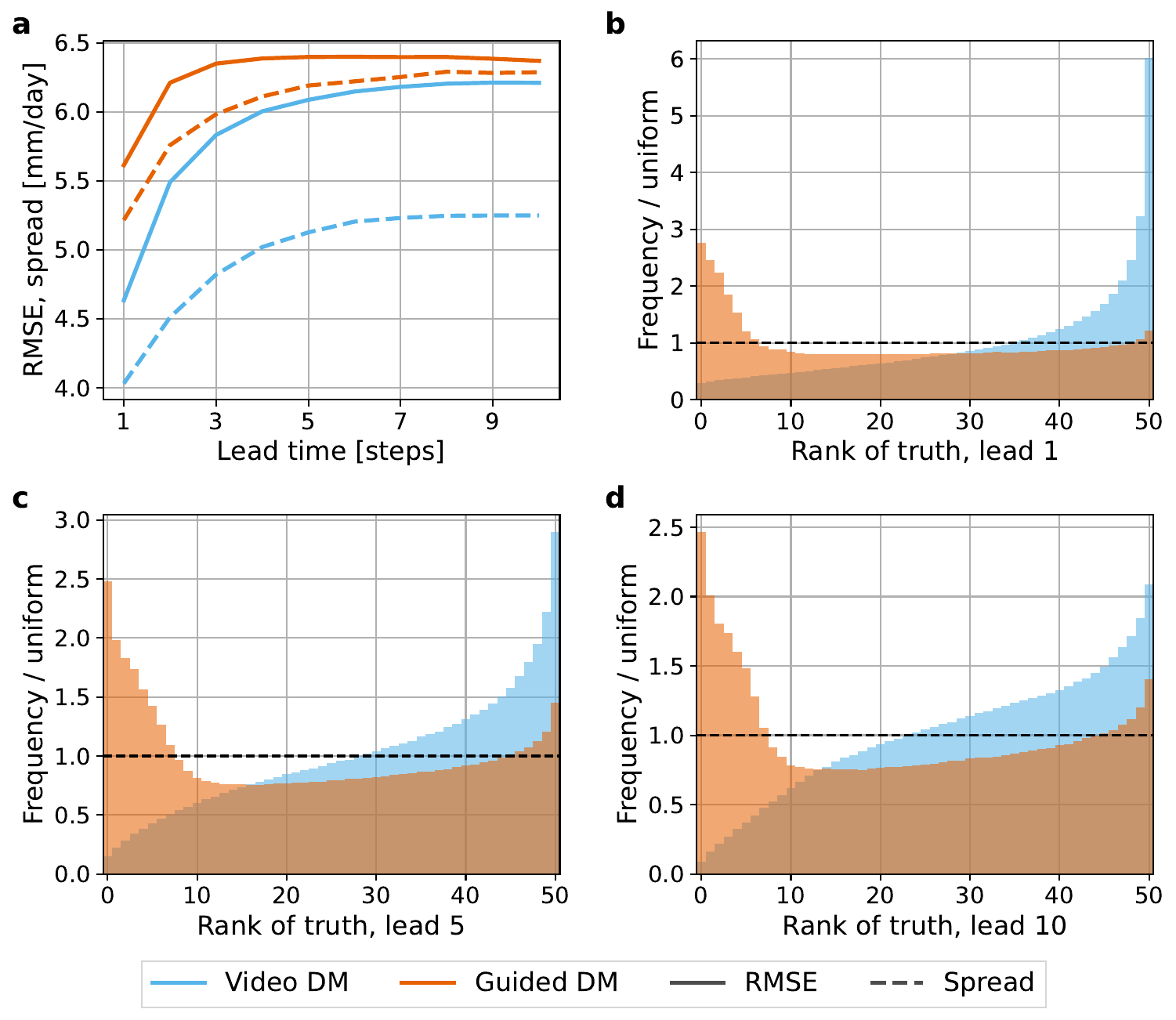}
    \caption{\textbf{Uncertainty calibration} of precipitation forecasts is shown for 100 ensemble forecasts with 50 members and 10-step lead time for the video DM (blue) and the guided DM (red). The RMSE (solid) and ensemble spread (dashed) are shown as a function of lead time (a), and rank histograms of the ground truth within the ensemble, normalized by the uniform distribution (black dashed), at lead times one (b), five (c) and ten (d).}
    \label{fig:calibration_precipitation}
\end{figure}

\begin{figure}[H]
    \centering
    \includegraphics[width=0.95\linewidth]{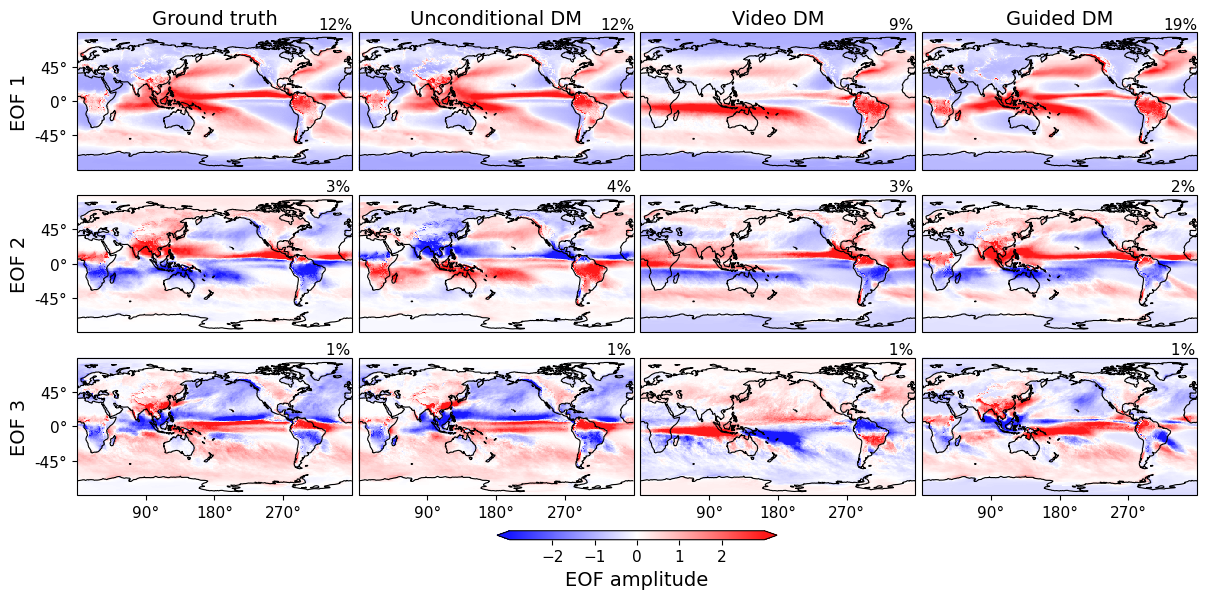}
    \caption{\textbf{Empirical orthogonal functions} (EOFs) are shown for the daily precipitation data for (from left to right) the ERA5 ground truth, the unconditional DM, the video DM and our guidance method. The explained variance is given in the top right of each panel. 
    }
    \label{fig:precipitation_eofs}
\end{figure}

\begin{figure}[H]
    \centering
    \includegraphics[width=0.8\linewidth]{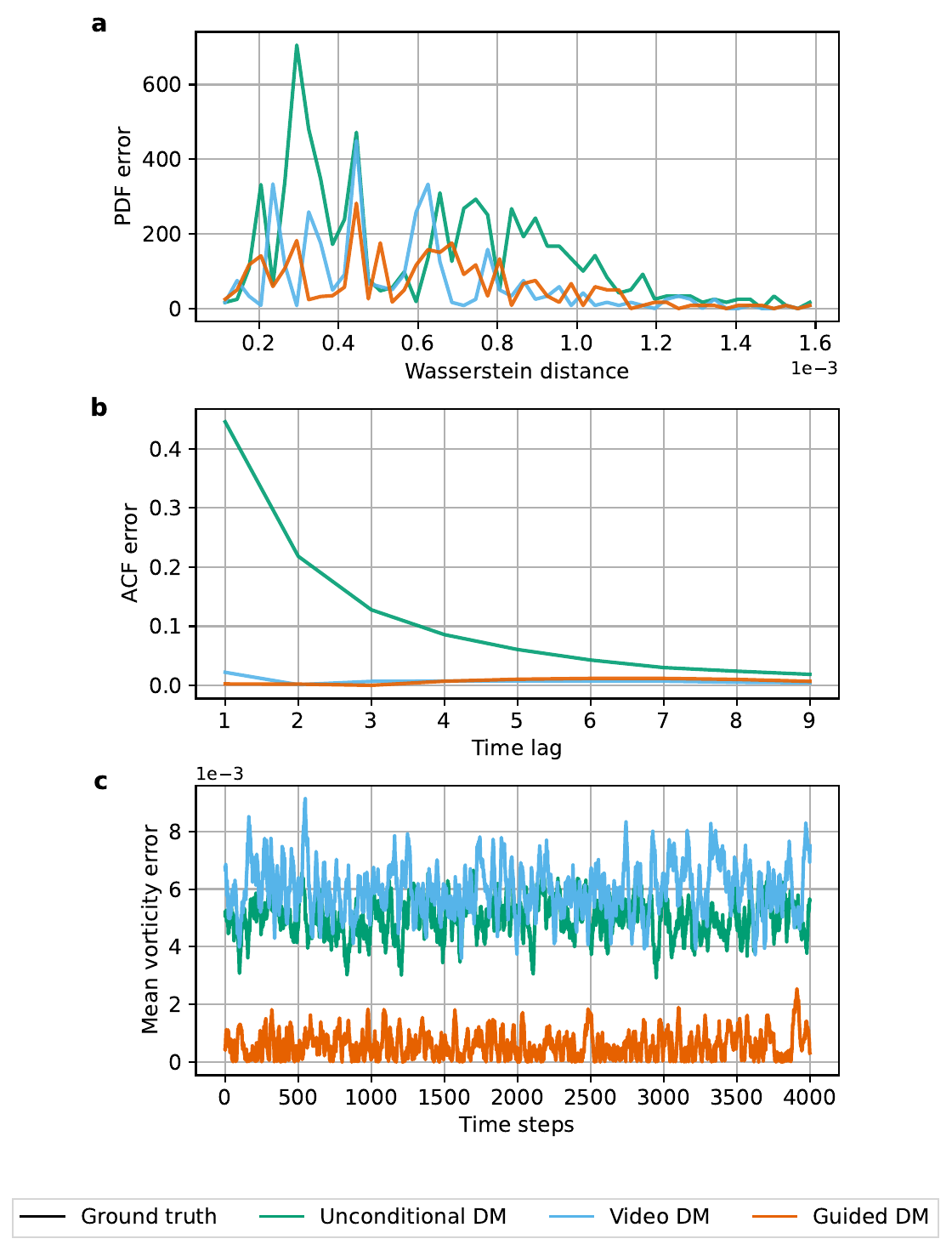}
    \caption{\textbf{Absolute errors of the vorticity statistics} shown in Fig.~\ref{fig:vorticity_statistics}, (a) Wasserstein-1 distance, the (b) autocorrelation functions (ACFs), and (c) the global average are shown for the unconditional DM (green), the video DM (blue) and our guidance method (red).}
    \label{fig:vorticity_stats_error}
\end{figure}

\begin{figure}[H]
    \centering
    \includegraphics[width=0.8\linewidth]{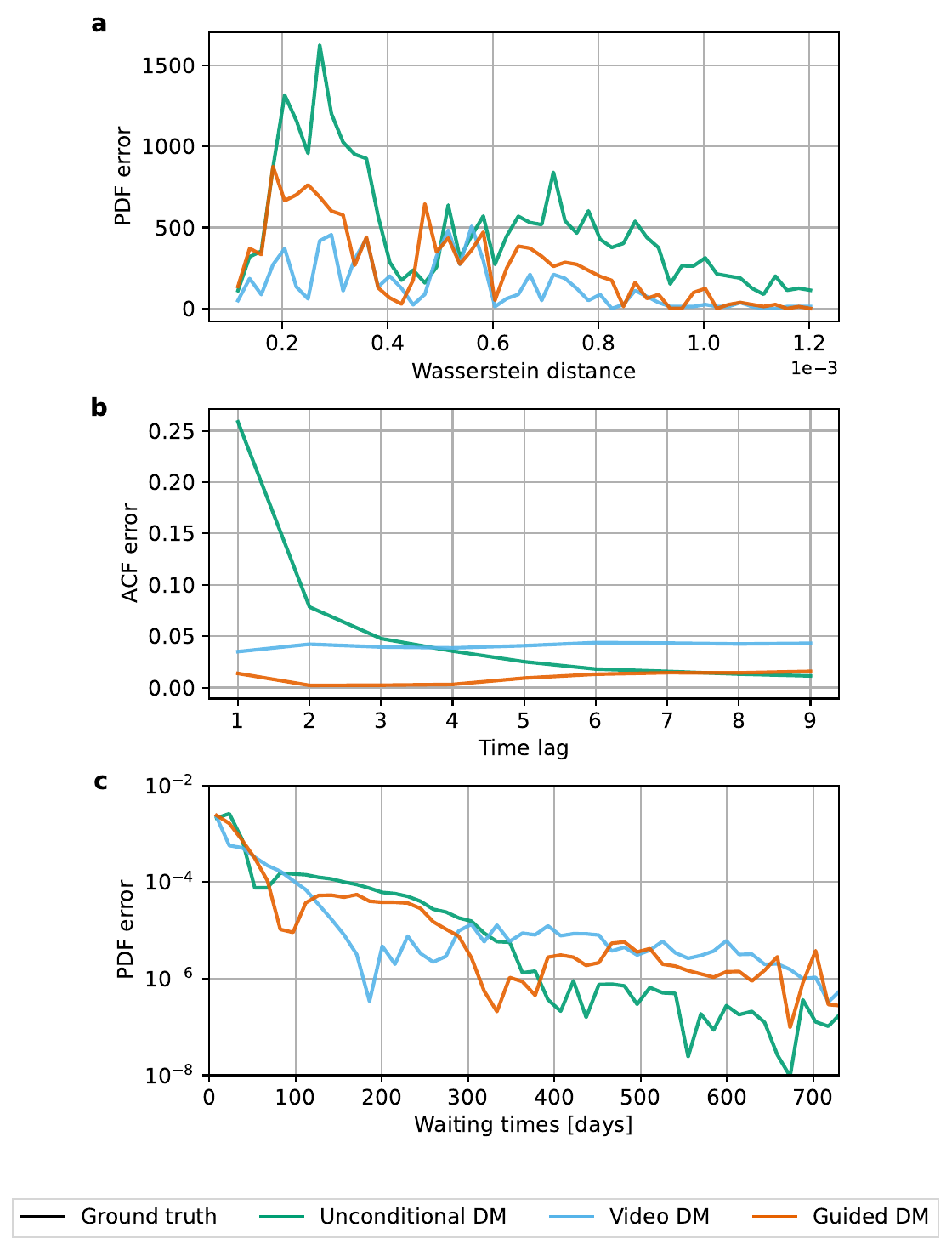}
    \caption{\textbf{Absolute errors of the precipitation statistics} shown in Fig.~\ref{fig:precipitation_statistics}, (a) Wasserstein-1 distance, the (b) autocorrelation functions (ACFs), and (c) waiting time distributions are shown for the unconditional DM (green), the video DM (blue) and our guidance method (red).}
    \label{fig:precipitation_stats_error}
\end{figure}

\subsection{Discriminator-guidance analysis}

\begin{figure}[H]
    \centering
    \includegraphics[width=1.0\linewidth]{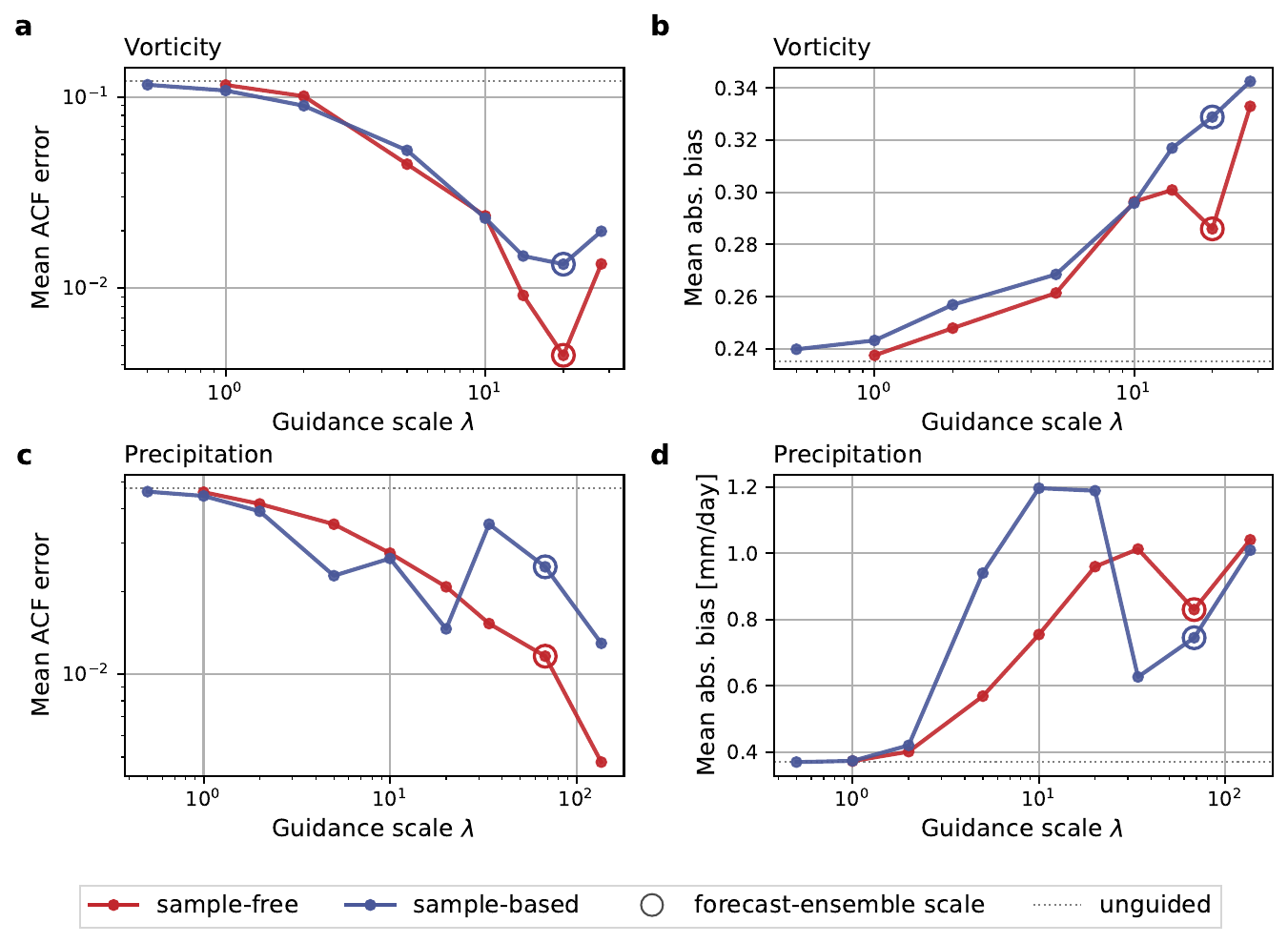}
    \caption{\textbf{Comparing guidance scales for sample-based and sample-free discriminators} in terms of the mean ACF error (a, c) and mean absolute bias (b, d) as a function of the guidance scale $\lambda$ for the vorticity (top) and precipitation (bottom) dataset, for the sample-free (red) and sample-based (blue) discriminator. Circles mark the scales used for the forecast ensembles in Fig.~\ref{fig:discriminator_comparison_vorticity} and \ref{fig:discriminator_comparison_precipitation}, and the dotted line shows the unguided DM.}
    \label{fig:discriminator_comparison_sweeps}
\end{figure}

\begin{figure}[H]
    \centering
    \includegraphics[width=1.0\linewidth]{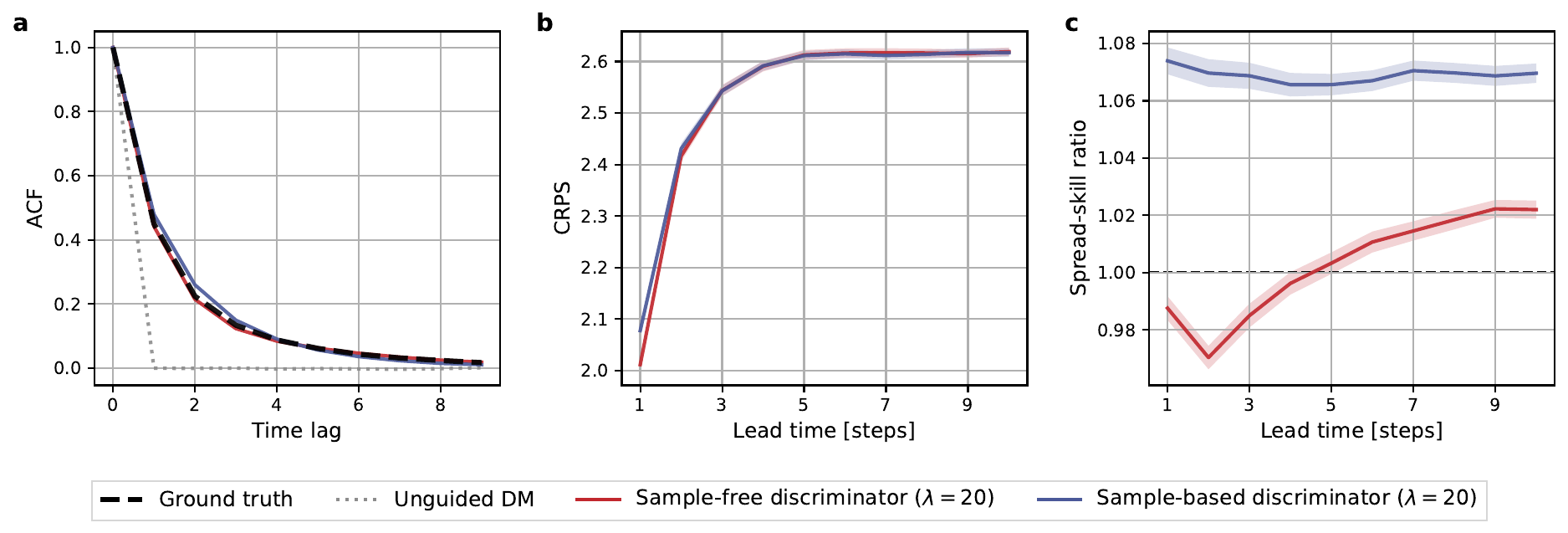}
    \caption{\textbf{Comparing time consistency for sample-based and sample-free discriminators on the vorticity dataset} in terms of ACFs (a), CRPS (b) and spread-skill ratio (c) for 100 ensemble forecasts with 50 members and 10-step lead time, for the ground truth simulation (black dashed), the unguided DM (grey dotted), the sample-free discriminator with $\lambda=20$ (red) and the sample-based discriminator with $\lambda=20$ (blue).}
    \label{fig:discriminator_comparison_vorticity}
\end{figure}

\begin{figure}[H]
    \centering
    \includegraphics[width=1.0\linewidth]{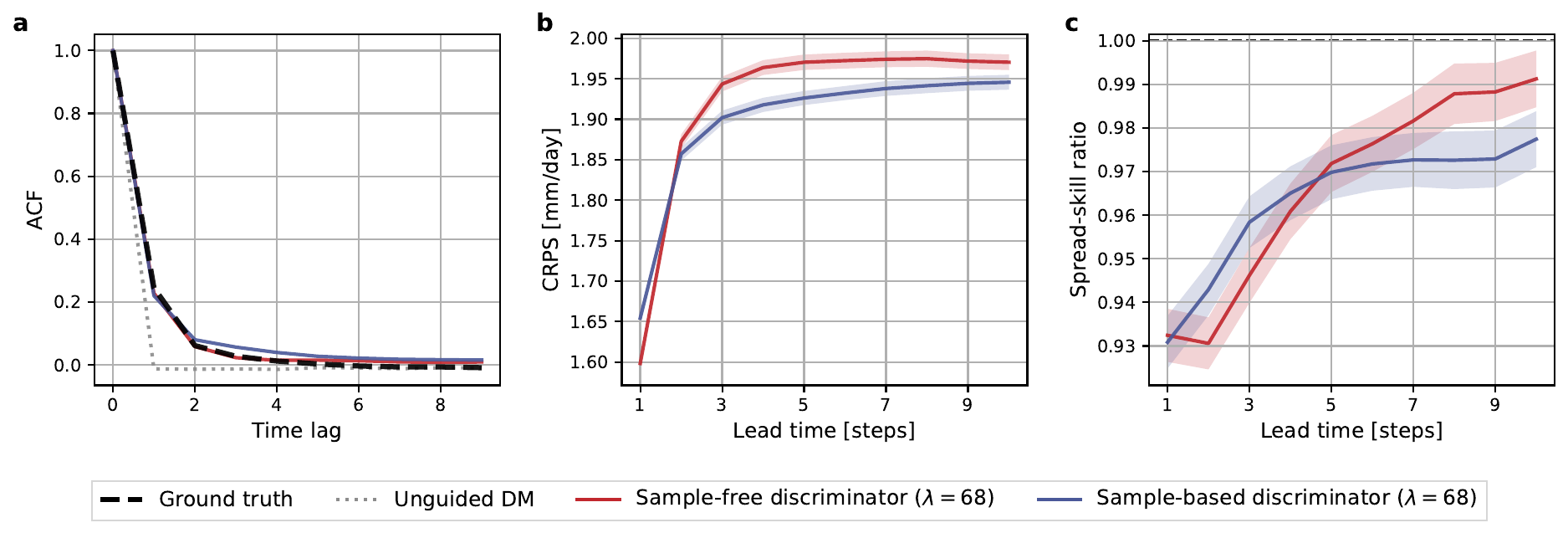}
    \caption{\textbf{Comparing time consistency for sample-based and sample-free discriminators on the precipitation dataset} in terms of ACFs (a), CRPS (b) and spread-skill ratio (c) for 100 ensemble forecasts with 50 members and 10-step lead time, for the ground truth ERA5 (black dashed), the unguided DM (grey dotted), the sample-free discriminator with $\lambda=68$ (red) and the sample-based discriminator with $\lambda=68$ (blue).}
    \label{fig:discriminator_comparison_precipitation}
\end{figure}

\begin{figure}[H]
    \centering
    \includegraphics[width=1.0\linewidth]{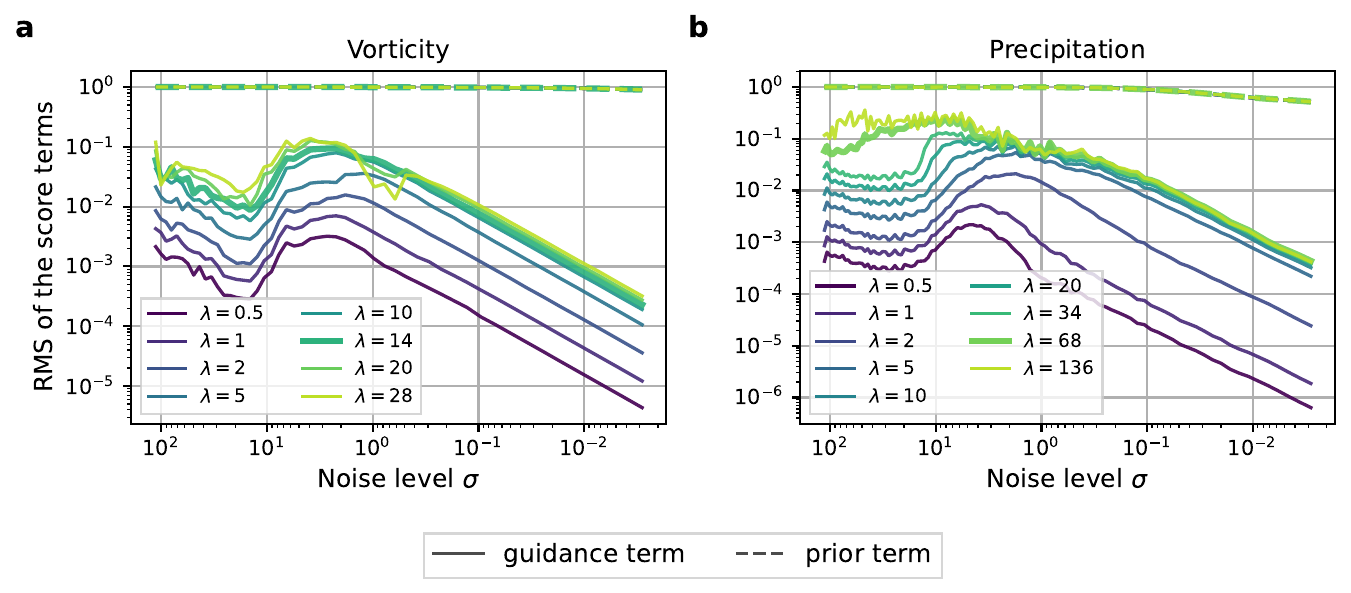}
    \caption{\textbf{Magnitude comparison of prior and guidance score} terms during sampling, shown as the root mean square (RMS) of the guidance term (solid) and the prior score term (dashed) as a function of the noise level $\sigma$ for different guidance scales $\lambda$ on the vorticity (a) and precipitation (b) dataset.}
    \label{fig:score_magnitudes}
\end{figure}

\begin{figure}[H]
    \centering
    \includegraphics[width=1.0\linewidth]{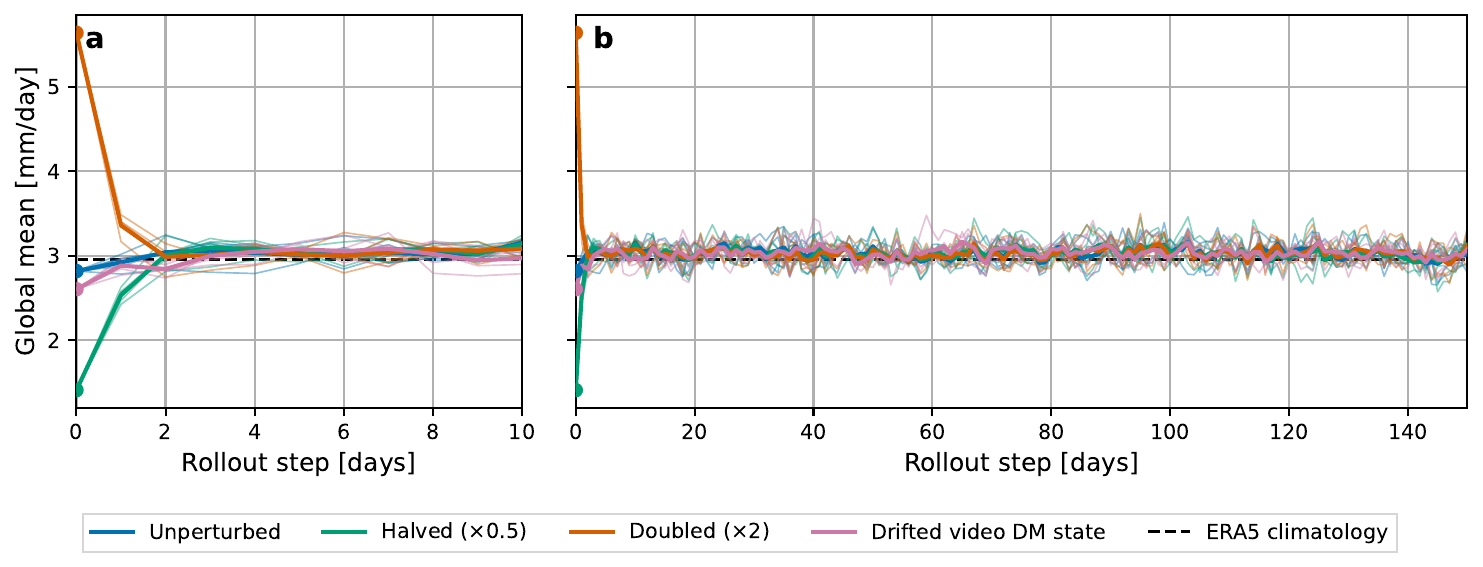}
    \caption{\textbf{Stability analysis} of the discriminator guidance with perturbed initial conditions. The global mean precipitation of five-member rollouts is shown for the first ten days (a) and the full 150 days (b), starting from the unperturbed test frames (blue), the test frames multiplied by 0.5 (green) and by 2 (orange), and from the end of the drifted video DM simulation in Fig.~\ref{fig:precipitation_100_years_forcing} (pink), together with the ERA5 climatology (black dashed).}
    \label{fig:restoring_force}
\end{figure}

\subsection{Generative adversarial network results}

\begin{figure}[H]
    \centering
    \includegraphics[width=1.0\linewidth]{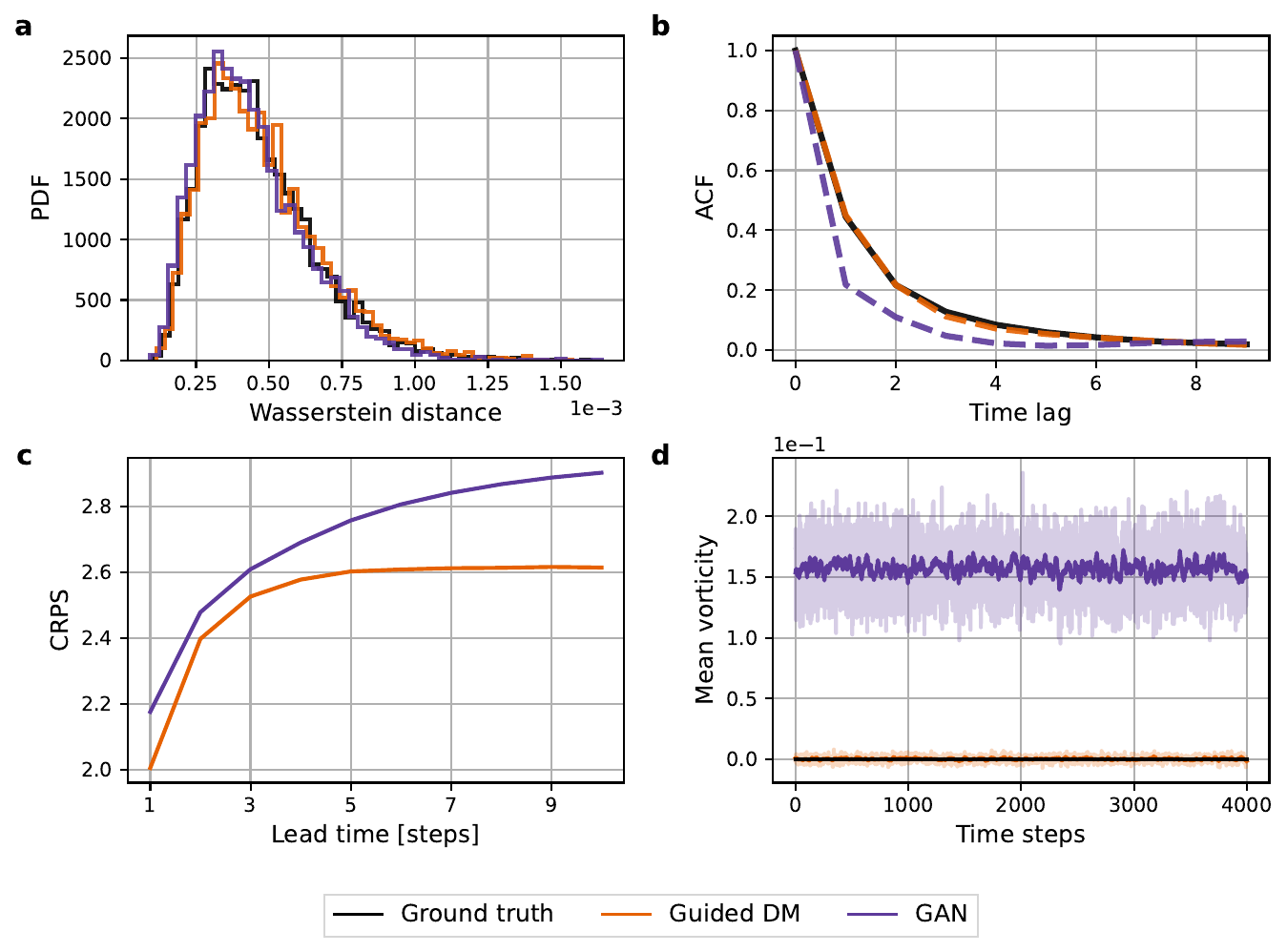}
    \caption{\textbf{Quantitative evaluation of the GAN baseline for 2D vorticity dynamics} in terms of Wasserstein distances between consecutive rows of the Hovmöller diagram in Fig.~\ref{fig:supplement_gan_vorticity_hovmoeller} (a), ACFs (b), CRPS (c) and running window spatial mean as solid line with the actual time series shown as shades (d), for the ground truth simulation (black), the guided DM (orange) and the GAN (purple).}
    \label{fig:supplement_gan_vorticity_stats}
\end{figure}

\begin{figure}[H]
    \centering
    \includegraphics[width=0.7\linewidth]{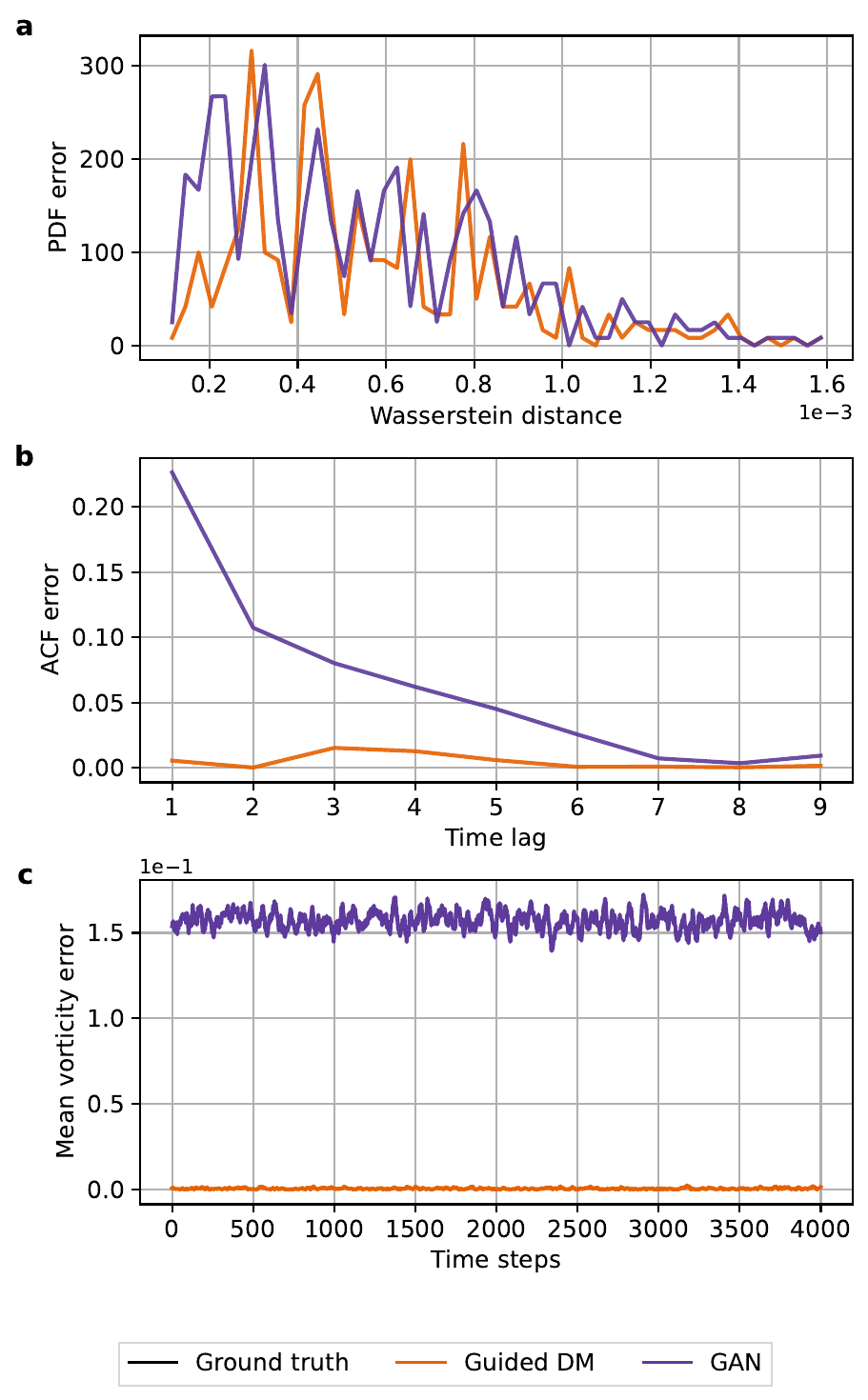}
    \caption{\textbf{Absolute errors of the GAN-based vorticity statistics} shown in Fig.~\ref{fig:supplement_gan_vorticity_stats}, (a) Wasserstein-1 distance, the (b) ACFs, and (c) the global average are shown for the guided DM (orange) and the GAN (purple).}
    \label{fig:supplement_gan_vorticity_stats_error}
\end{figure}

\begin{figure}[H]
    \centering
    \includegraphics[width=0.7\linewidth]{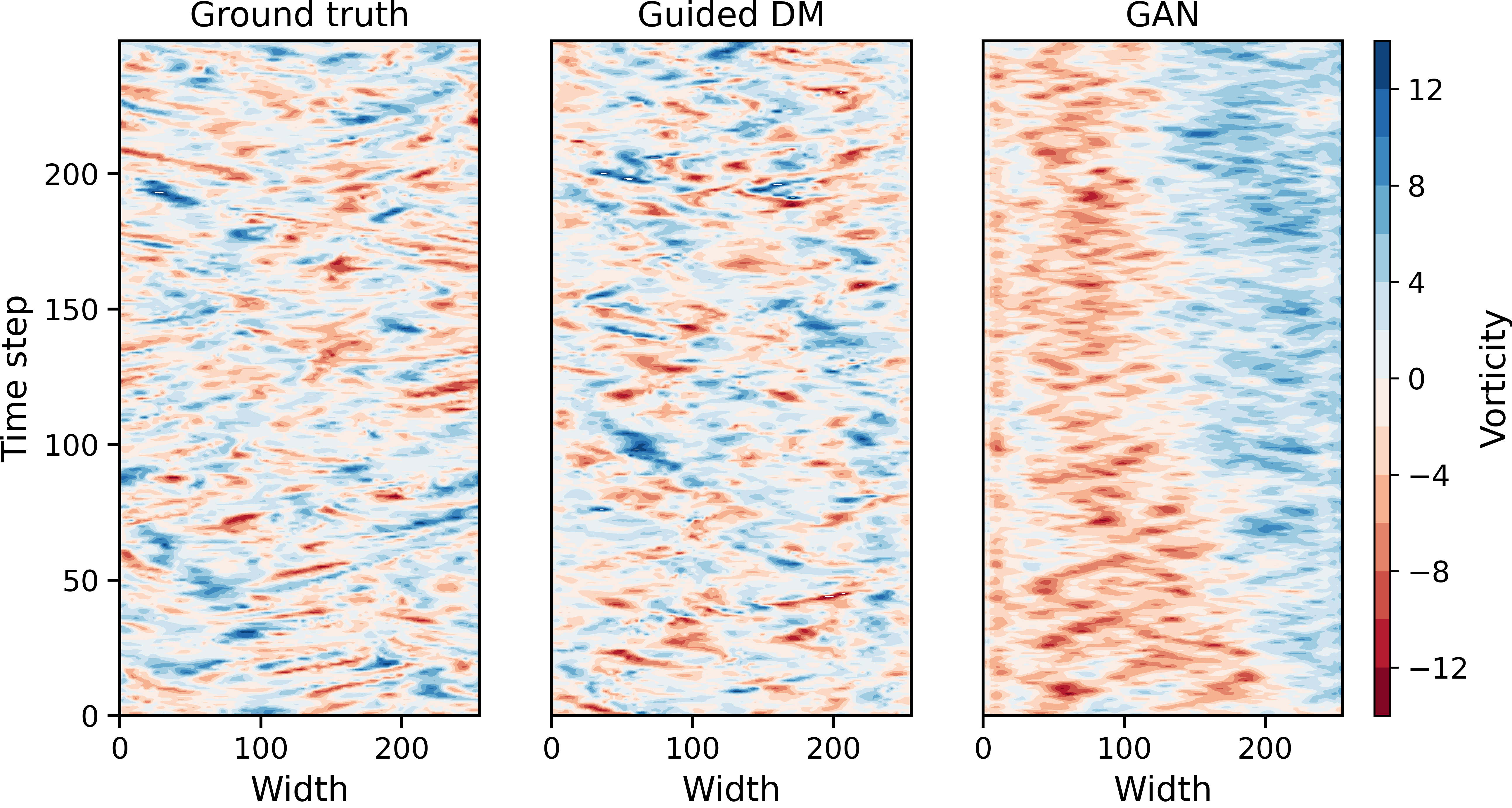}
    \caption{\textbf{Hovmöller diagrams} of the 2D vorticity simulation are computed as in Fig.~\ref{fig:vorticity_hovmoeller} for (from left to right) the ground truth numerical simulation, the guided DM and the GAN.}
    \label{fig:supplement_gan_vorticity_hovmoeller}
\end{figure}

\begin{figure}[H]
    \centering
    \includegraphics[width=1.0\linewidth]{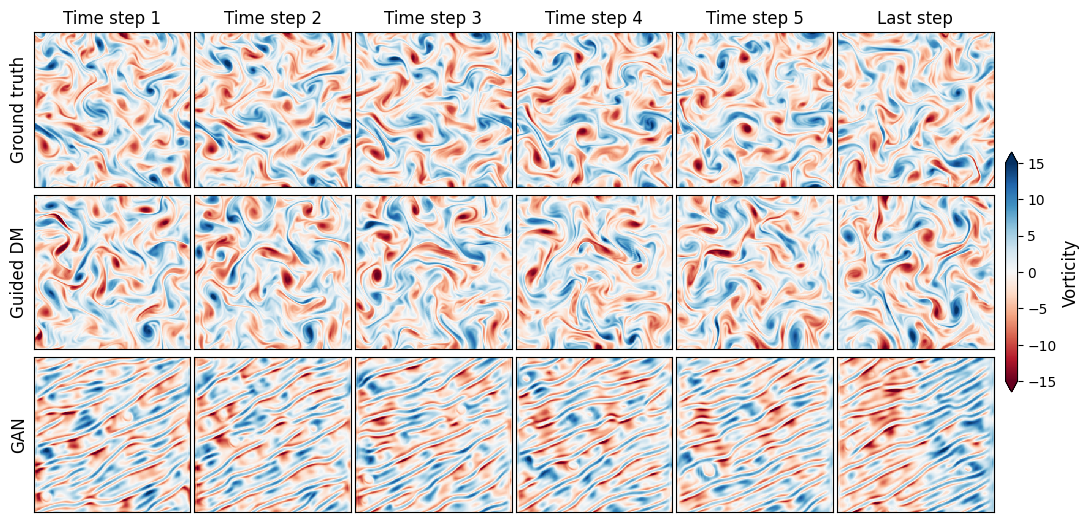}
    \caption{\textbf{Qualitative comparison} of the first five and last 2D vorticity fields from the direct numerical Navier-Stokes turbulence simulation (top), the guided DM (middle) and the GAN (bottom). Each row shows a single rollout starting from the same initial condition.}
    \label{fig:supplement_gan_vorticity_samples}
\end{figure}

\begin{figure}[H]
    \centering
    \includegraphics[width=0.55\linewidth]{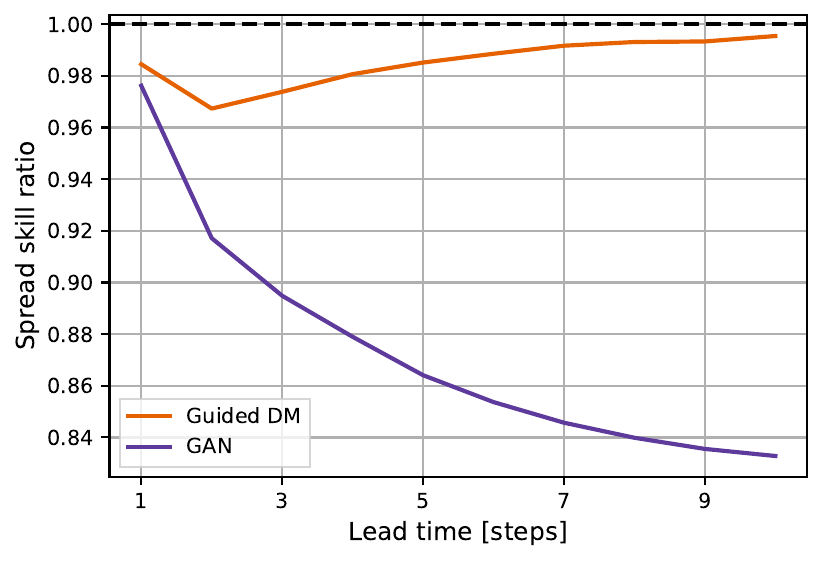}
    \caption{\textbf{The spread skill ratio} (SSR) of vorticity forecasts is shown for 100 ensemble forecasts with 50 members and 10-step lead time for the (orange) guided DM and (purple) GAN. A perfect forecast would have a SSR of one.}
    \label{fig:supplement_gan_vorticity_spread_skill_ratio}
\end{figure}

\begin{figure}[H]
    \centering
    \includegraphics[width=1.0\linewidth]{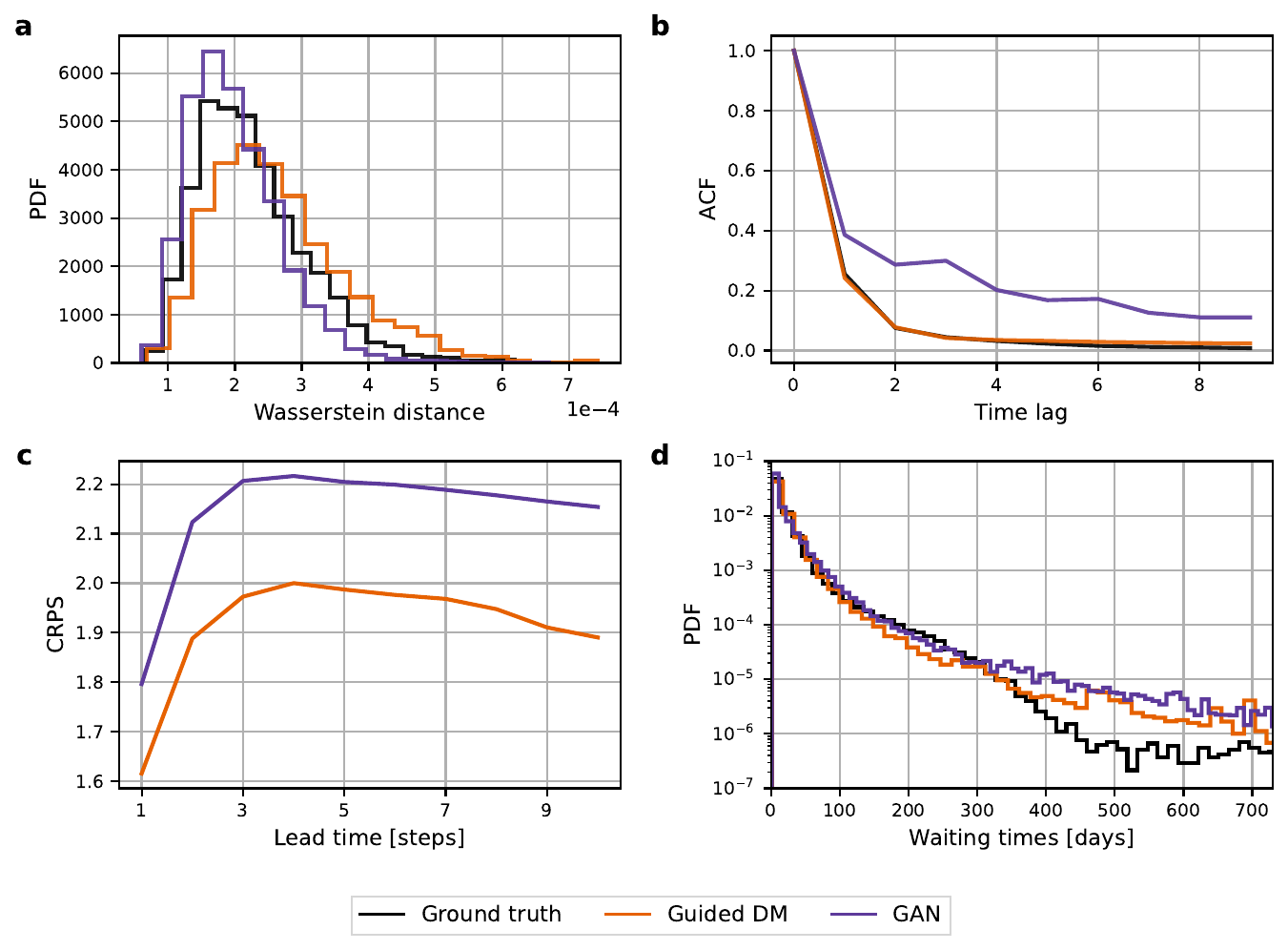}
    \caption{\textbf{Quantitative evaluation of the GAN baseline for daily precipitation dynamics} in terms of Wasserstein distances between consecutive rows of the Hovmöller diagram in Fig.~\ref{fig:supplement_gan_precipitation_hovmoeller} (a), ACFs (b), CRPS forecast skill (c), extreme event waiting time distributions (d), for the ground truth ERA5 (black), the guided DM (orange) and the GAN (purple).}
    \label{fig:supplement_gan_precipitation_stats}
\end{figure}

\begin{figure}[H]
    \centering
    \includegraphics[width=0.7\linewidth]{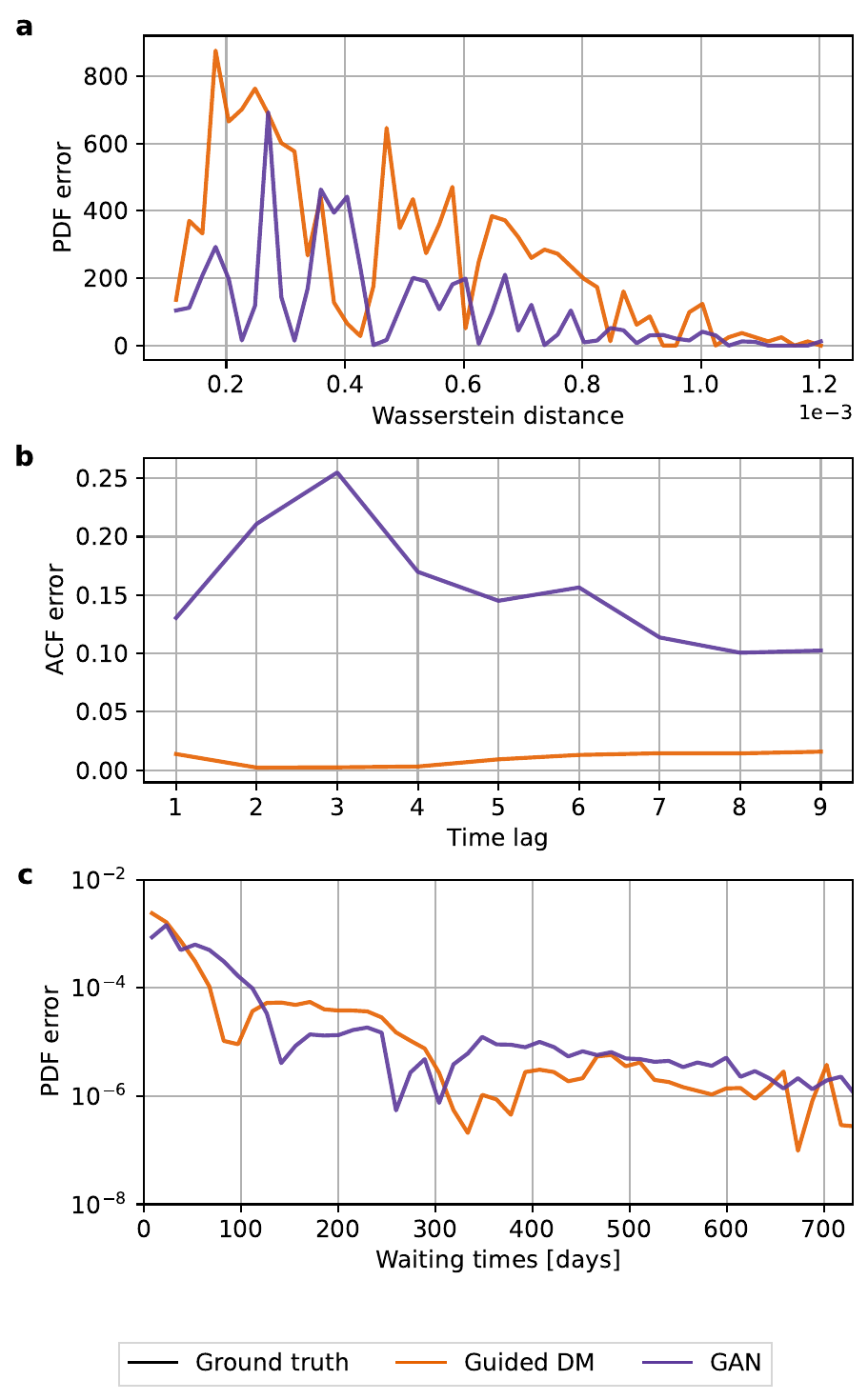}
    \caption{\textbf{Absolute errors of the GAN-based precipitation statistics} shown in Fig.~\ref{fig:supplement_gan_precipitation_stats}, (a) Wasserstein-1 distance, the (b) ACFs, and (c) waiting time distributions are shown for the guided DM (orange) and the GAN (purple).}
    \label{fig:supplement_gan_precipitation_stats_error}
\end{figure}

\begin{figure}[H]
    \centering
    \includegraphics[width=0.7\linewidth]{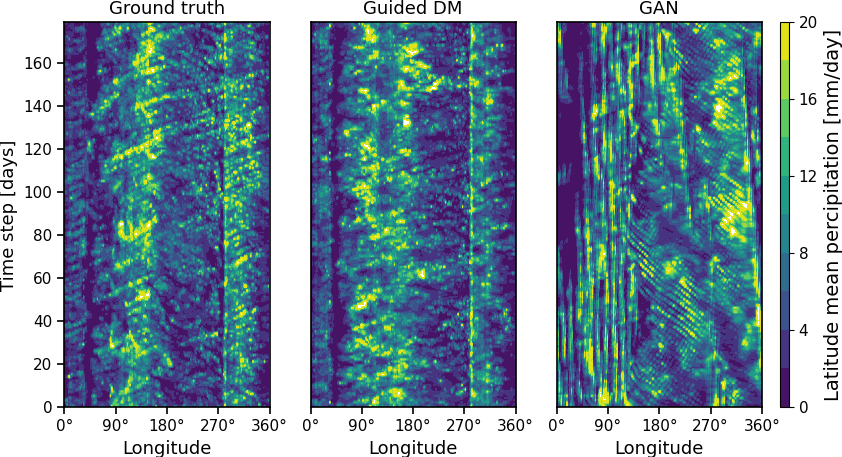}
    \caption{\textbf{Hovmöller diagrams} of the global daily precipitation simulation (from left to right) are computed for 180 days as a mean over the latitude band from $10^\circ$S to $10^\circ$N for the ground truth ERA5, the guided DM and the GAN.}
    \label{fig:supplement_gan_precipitation_hovmoeller}
\end{figure}

\begin{figure}[H]
    \centering
    \includegraphics[width=1.0\linewidth]{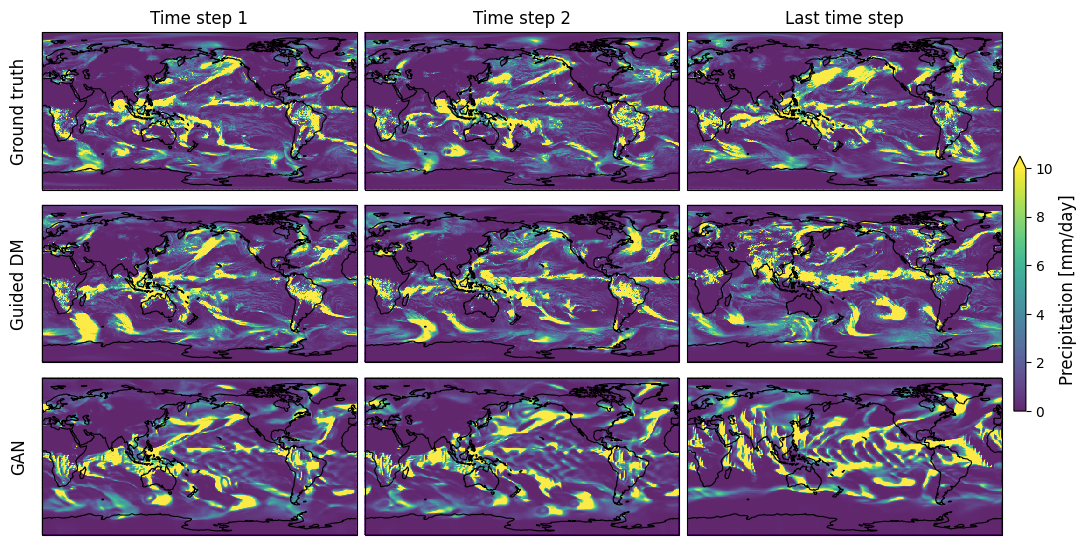}
    \caption{\textbf{Qualitative comparison} of the first two and last daily precipitation fields from the ERA5 ground truth (top), the guided DM (middle) and the GAN (bottom).}
    \label{fig:supplement_gan_precipitation_samples}
\end{figure}

\begin{figure}[H]
    \centering
    \includegraphics[width=0.55\linewidth]{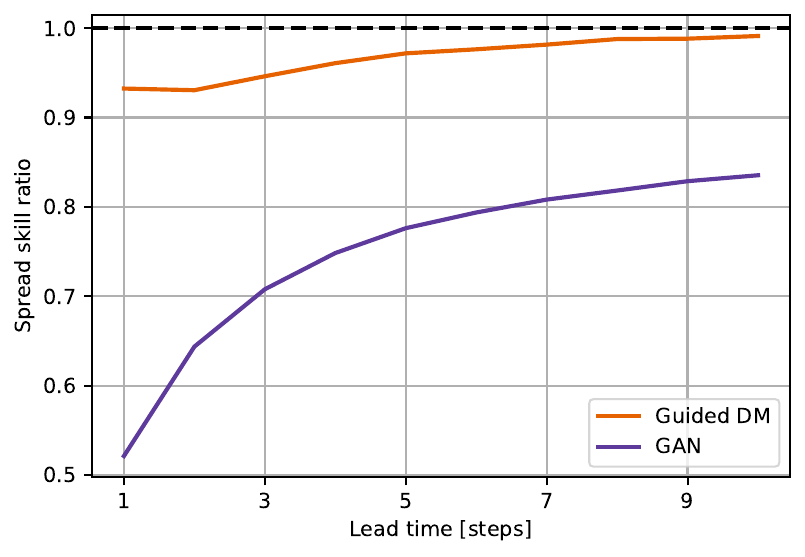}
    \caption{\textbf{The spread skill ratio} (SSR) of precipitation forecasts is shown for 100 ensemble forecasts with 50 members and 10-step lead time for the (orange) guided DM and (purple) GAN.}
    \label{fig:supplement_gan_precipitation_spread_skill_ratio}
\end{figure}

\begin{figure}[H]
    \centering
    \includegraphics[width=1.0\linewidth]{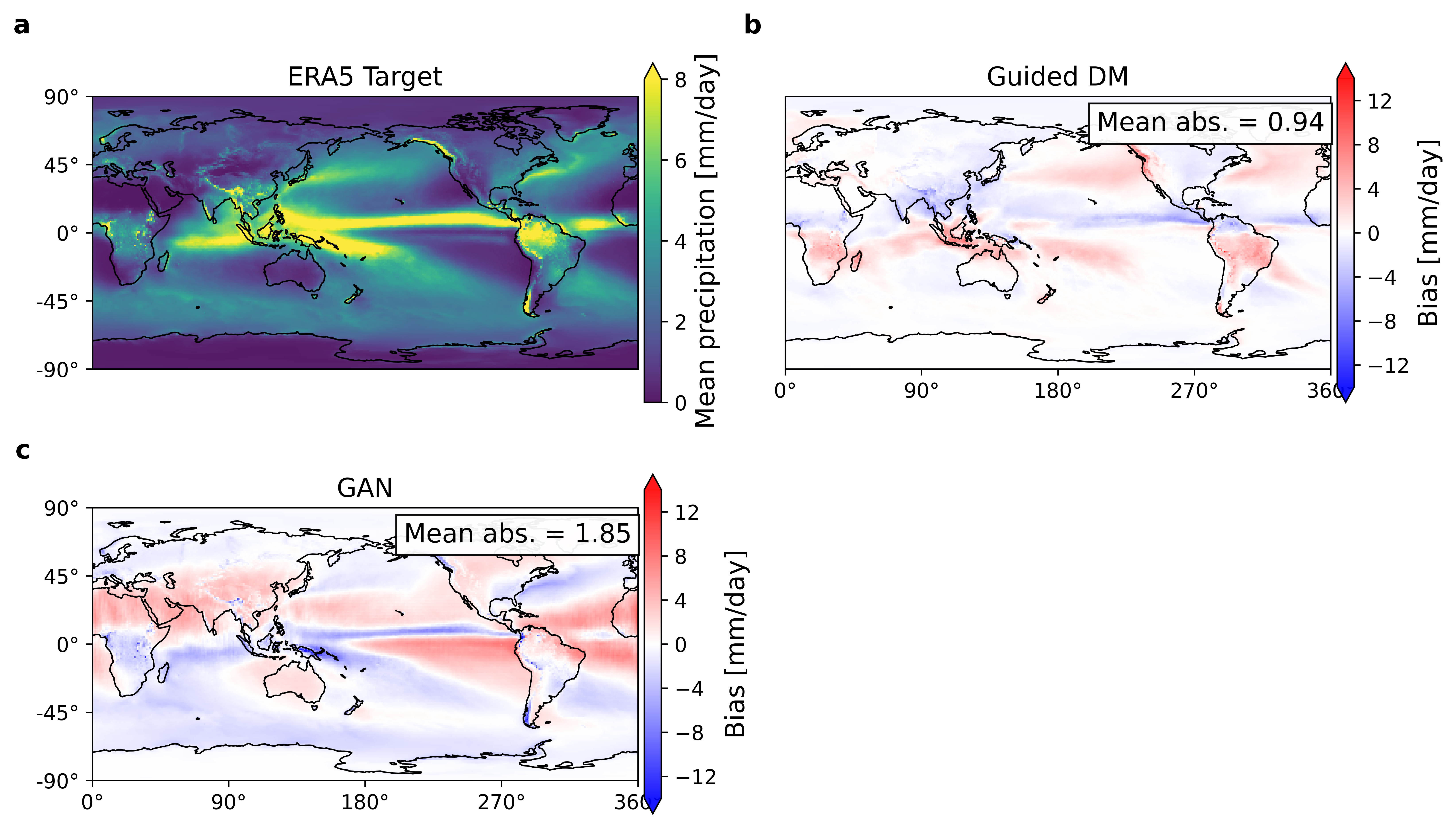}
    \caption{\textbf{Global mean bias} (see Appendix \ref{app:metrics} for definition) of the GAN baseline showing, (a) the test set mean of the ERA5 ground truth, (b) the bias of the guided DM, (c) the GAN. Mean absolute bias values are given in the top right.}
    \label{fig:supplement_gan_precipitation_bias}
\end{figure}

\vskip 0.2in
\bibliography{references}

\end{document}